\documentclass[10pt,twocolumn,letterpaper]{article}

\usepackage[pagenumbers]{cvpr}

\usepackage{graphicx}        
\usepackage{booktabs}        
\usepackage{tcolorbox}       
\tcbuselibrary{skins,breakable}  

\usepackage[utf8]{inputenc}
\usepackage[T1]{fontenc}
\usepackage{url}
\usepackage{amsfonts}
\usepackage{nicefrac}
\usepackage{microtype}
\usepackage{placeins}
\usepackage{adjustbox}
\usepackage{comment}
\usepackage{arydshln}
\usepackage{xspace}
\usepackage{float}
\usepackage{enumitem}
\usepackage{wrapfig}
\usepackage{makecell}
\usepackage{todonotes}
\usepackage{tabularx}
\usepackage{graphicx}
\usepackage{fontawesome6}    

\definecolor{cvprblue}{rgb}{0.21,0.49,0.74}
\usepackage[pagebackref,breaklinks,colorlinks,allcolors=cvprblue]{hyperref}

\def\paperID{*****} 
\def\confName{CVPR}
\def\confYear{2026}

\title{%
  \makebox[0pt][r]{\raisebox{-0.3\height}{\includegraphics[height=3em]{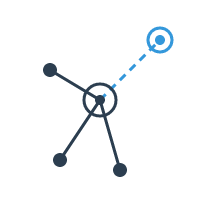}}}%
  Communicating about Space:\\
  Language-Mediated Spatial Integration Across Partial Views
}

\author{
Ankur Sikarwar$^{1,2}$\thanks{Equal contribution} \quad
Debangan Mishra$^{3}$\footnotemark[1] \quad
Sudarshan Nikhil$^{3}$ \\[0.3em]
Ponnurangam Kumaraguru$^{3}$ \quad
Aishwarya Agrawal$^{1,2,4}$ \\[0.5em]
$^{1}$Mila -- Quebec AI Institute \quad $^{2}$Université de Montréal \quad $^{3}$IIIT Hyderabad \\ $^{4}$Canada CIFAR AI Chair
}

\begin{document}
\maketitle
\begin{abstract}
Humans build shared spatial understanding by communicating partial, viewpoint-dependent observations. We ask whether Multimodal Large Language Models (MLLMs) can do the same, aligning distinct egocentric views through dialogue to form a coherent, allocentric mental model of a shared environment. To study this systematically, we introduce \textbf{\textsc{Cosmic}}, a benchmark for \emph{\textbf{Co}llaborative \textbf{S}patial Co\textbf{m}mun\textbf{ic}ation}. In this setting, two static MLLM agents observe a 3D indoor environment from different viewpoints and exchange natural-language messages to solve spatial queries.
\textsc{Cosmic} contains 899 diverse scenes and 1{,}250 question--answer pairs spanning five tasks. We find a
capability hierarchy, MLLMs are most reliable at identifying shared anchor objects across views, perform worse on relational reasoning, and largely fail at building globally consistent maps, performing near chance, even for frontier models. Moreover, we find thinking capability yields gains in anchor grounding, but is insufficient for higher-level spatial communication.
To contextualize model behavior, we collect 250 human--human dialogues. Humans achieve 95\% aggregate accuracy, while the best model, Gemini-3-Pro-Thinking, reaches 72\%, leaving substantial room for improvement.
Moreover, human conversations grow more precise as partners align on a shared spatial understanding, whereas MLLMs keep exploring without converging, suggesting limited capacity to form and sustain a robust shared mental model throughout the dialogue.
Our code and data is available at \href{https://github.com/ankursikarwar/Cosmic}{link}.

\end{abstract}



\begin{figure*}[tb]
  \centering
  \includegraphics[width=0.925\textwidth]{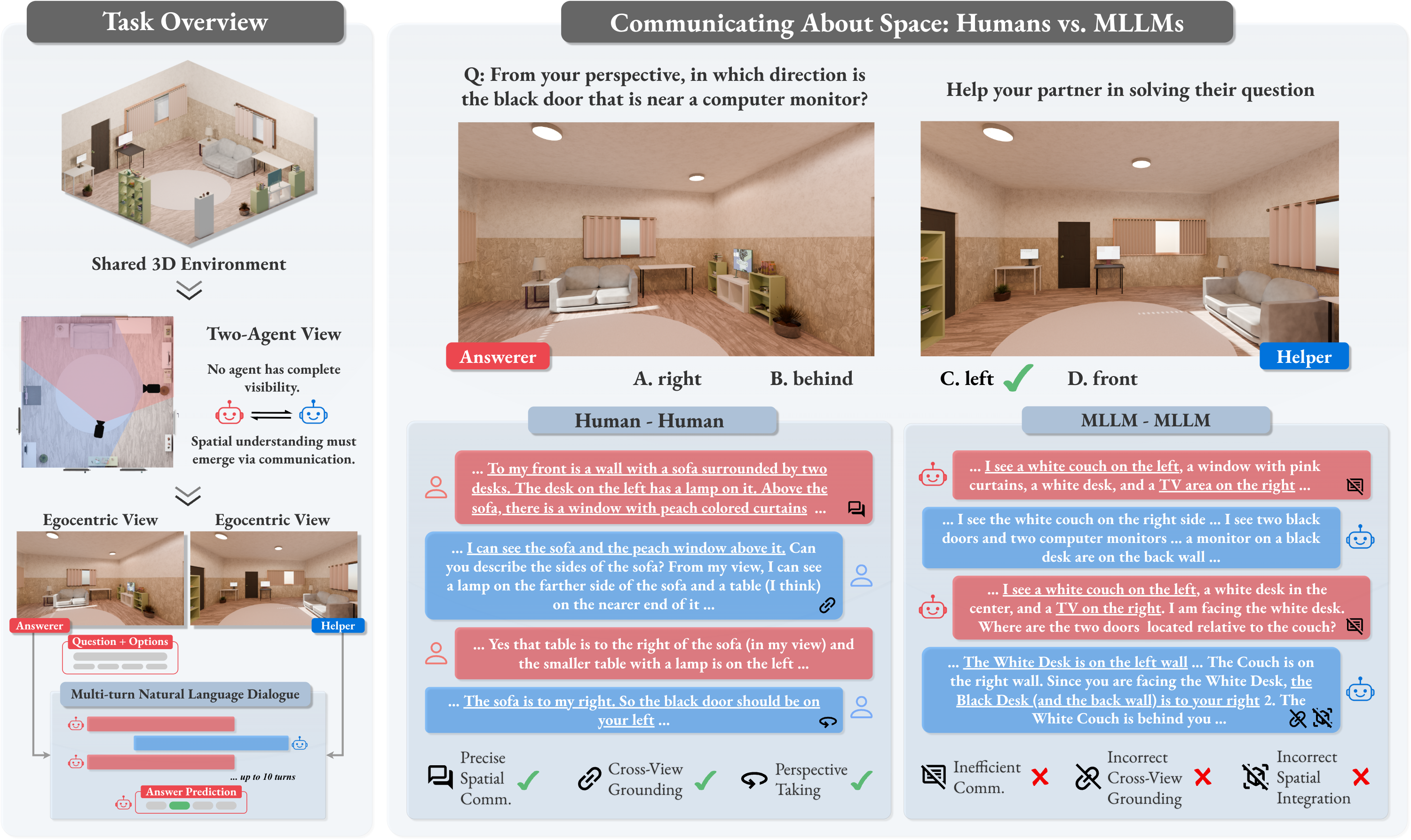}
  \caption{
      \textbf{Left:} MLLM agents attempt to communicate and build a spatial mental model for answering questions in \textsc{COSMIC}. \textbf{Right:} Answerer and Helper agents integrate distinct egocentric views via communication to answer the question. Humans demonstrate efficient and precise strategies while MLLM agents are more verbose, inefficient, and fail to build and maintain a robust shared mental model. 
    }
    \vspace{-4mm}
  \label{fig:teaserr}
\end{figure*}

\section{Introduction}
\label{sec:intro}

Spatial intelligence extends beyond individual perception. Through communication and collaboration, humans transform local observations into shared spatial mental models \cite{Levinson2003Space, Tversky2003SpatialSchemas, Clark1996UsingLanguage, CognitiveMaps1, garrod1987saying}.
Imagine two friends trying to meet at a large park they have never visited. One says, \textit{``I am near the lamp post beside the fountain,''} and the other replies, \textit{``I see a tall tree and the fountain.''} Neither has a complete view, but through dialogue they integrate these utterances into a shared spatial model. They ground common landmarks (\textit{fountain}), infer relative orientation, and progressively refine hypotheses about unseen regions. Language here acts as a scaffold for spatial communication \cite{Clark1996UsingLanguage, Levinson2003Space}, enabling behaviors like frame-of-reference switching, grounding shared objects across views (\emph{anchor objects}), clarification and repair, and the overall synthesis of complementary partial views. 

This ability to build shared spatial models through communication is increasingly central for modern AI systems. As MLLMs are now expected to operate as collaborative partners rather than passive tools, deployed across embodied robotics~\cite{tellex2011understanding}, AR/VR platforms, and multi-user interactive systems~\cite{park2023generative}, spatial reasoning becomes inherently distributed, with no single agent having access to the full environment. Unlike single-agent settings \cite{mmsi, seeingfromanother, ThinkingInSpace}, these distributed scenarios require agents to communicate observations, reconcile conflicting interpretations, and build a consistent shared model of the environment, all through natural language. Yet despite rapid progress on spatial perception and reasoning benchmarks \cite{ThinkingInSpace, viewspatial, mincube, seeingfromanother}, the capacity of MLLMs for this kind of communication-driven spatial integration remains largely unexplored.

We address this gap with \textbf{\textsc{Cosmic}}, a diagnostic benchmark designed to test whether MLLM agents can build a shared spatial understanding through dialogue. The benchmark places two agents in the same indoor scene from distinct egocentric perspectives, requiring
multi-turn dialogue to integrate complementary observations and jointly solve a QA task (Fig.~\ref{fig:teaserr}). Our key contributions are as follows:
\begin{enumerate}
    \item We introduce the task of building shared spatial understanding through natural-language dialogue across partial, egocentric views. To support systematic evaluation of this capability, we present \textbf{\textsc{Cosmic}}, a \emph{diagnostic} benchmark comprising 899 procedurally generated diverse indoor scenes and 1,250 question-answer pairs spanning five tasks evaluating object-level, relation-level, and map-level spatial reasoning capabilities.
    \item We conduct zero-shot evaluations of state-of-the-art open-source and proprietary MLLMs, revealing a deterioration in their capabilities from anchor grounding to relational reasoning to cognitive mapping. Through systematic failure mode analysis, we further demonstrate that cross-view grounding and geometric reasoning constitute the primary limitations in current MLLMs.
    \item We collect 250 human--human dialogues and provide a systematic comparison of human and model dialogues in collaborative spatial reasoning, revealing that humans converge rapidly through targeted, information-dense exchanges grounded in precise spatial descriptions, while MLLM agents resort to verbose, exploratory dialogue that remains shallow and fails to build a coherent shared spatial model (see Fig. \ref{fig:teaserr} for an example). 

\end{enumerate}
\begin{figure*}[tb]
  \centering
  \includegraphics[width=0.94\textwidth]{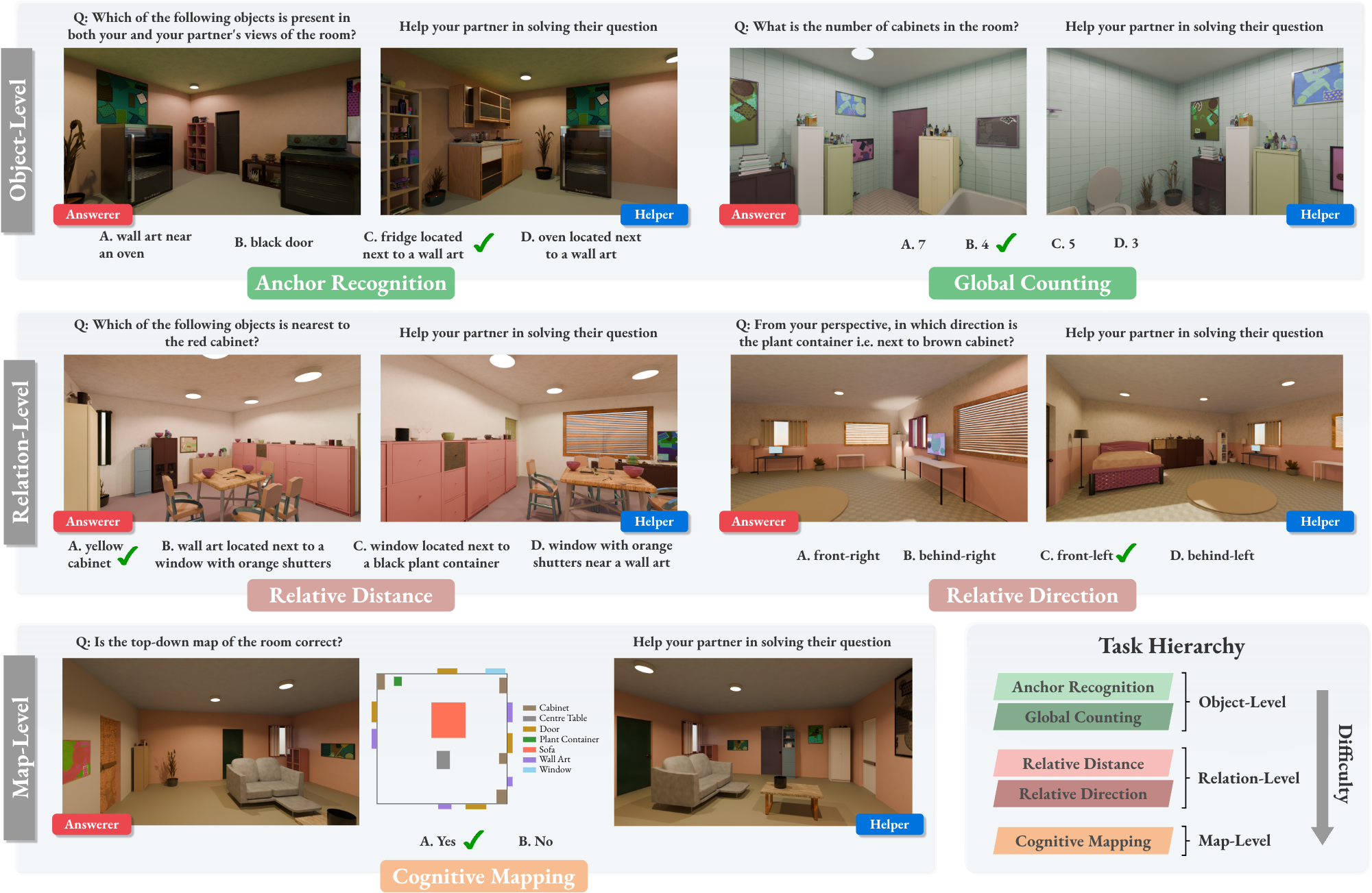}
  \caption{
      \textbf{Overview of the \textsc{COSMIC} benchmark.} Each task pair shows the Answerer's view (left) and the Helper's view (right), along with the question and options posed to the Answerer.  
    }
    \vspace{-4mm}
  \label{fig:benchmark}
\end{figure*}

\section{Related Work}
\label{sec:related_work}

\noindent\textbf{Single-view Spatial reasoning.} Early studies revealed that vision--language models such as CLIP and BLIP-VQA struggle with even basic spatial relationships \cite{kamath2023whatsup}. Subsequent work systematically evaluated MLLMs on spatial reasoning, with benchmarks such as What'sUp \cite{kamath2023whatsup} and VSR \cite{liu2023visualspatialreasoning} probing basic positional understanding, and SpatialRGPT-Bench \cite{cheng2024spatialrgptgroundedspatialreasoning} extending evaluation to broader aspects of spatial cognition. More recent benchmarks, including OmniSpatial \cite{jia2026omnispatialcomprehensivespatialreasoning} and \cite{stogiannidis2025mindgapbenchmarkingspatial}, further expand the scope to complex real-world scenarios and tasks requiring mental rotation, folding, and navigation. However, these benchmarks are limited to reasoning over single-view.

\noindent\textbf{Multi-view spatial reasoning.} Recent benchmarks extend spatial evaluation beyond single views by presenting multiple images of a scene to a single model \cite{li2025viewspatial, ThinkingInSpace, lee2025spatialmosaic}. \textsc{MindCube} \cite{yin2025mindcube} probes spatial mental models through questions involving translations, rotations, and perspective-taking, while Ego3D-Bench \cite{gholami2025spatialreasoningvisionlanguagemodels} and All-Angles Bench \cite{seeingfromanother} evaluate models on egocentric multi-view observations. Fine-tuning with reinforcement learning or supervised objectives can partially mitigate model's limitations \cite{yin2025mindcube, seeingfromanother} on spatial reasoning. All of these benchmarks remain centralized, with a single model reasoning over all available views. 
\textsc{Cosmic} addresses the fundamentally harder distributed setting where \emph{two agents}, each with partial observations, are required to reason jointly.

\noindent\textbf{Multi-agent cooperation and communication.} Prior work has examined multi-agent LLM and MLLM systems on collaborative tasks. Benchmarks such as \cite{badola2025MultiTurnPuzzles, li2025hiddenbench} evaluate how LLMs handle missing information through sustained dialogue to solve textual puzzles. In \cite{ossowski2024comma, xu2025vsbench}, multimodal agents with complementary observations cooperate to solve games, while \cite{zhang2025mmcnav, zhu2026cavln} explore similar settings in navigation. Other works propose frameworks for improving reasoning and assessing collaborative behavior in communication-based environments \cite{chen2024reconcile, chen2024comm, du2024MAD}. However, multi-agent spatial reasoning under egocentric partial observability remains understudied.

\noindent\textbf{Cognitive maps.} Cognitive science conceptualizes scene understanding through \emph{cognitive maps}, internal representations encoding environmental geometry \cite{tolman1948cognitivemaps, o1978hippocampus}. Recent work investigates whether MLLMs can form such map-like representations \cite{ThinkingInSpace, yin2025mindcube, gao2026map2thoughtexplicit3dspatial}. 
Unlike humans, who routinely build such representations collaboratively from partial observations, these works focus on centralized, single-model settings. 
\textsc{Cosmic} bridges this gap by studying map-level understanding in a distributed, communication-based multi-agent setting.
\begin{figure*}
\includegraphics[width=\linewidth]{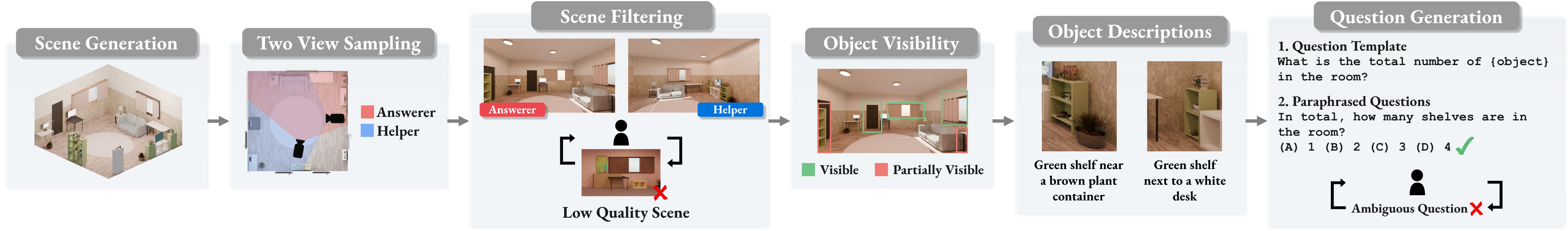}
    \caption{\textbf{Benchmark curation pipeline.} Our pipeline involves scene generation, sampling complementary agent viewpoints, generating questions using templates and unique object description followed by paraphrasing the questions.}
    \vspace{-5.5mm}
    \label{fig:pipeline}
\end{figure*}

\section{\textsc{Cosmic} Benchmark}
\label{sec:data}

\subsection{Overview}

Our task involves two static MLLM-based agents, an Answerer ($A$) and a Helper ($H$), positioned at different locations within a 3D indoor room environment. Each agent receives their own egocentric RGB view ($I_{A}$ \& $I_{H}$), providing them with partial but complementary perspectives of the environment. The Answerer is additionally given a question and is required to collaborate with the Helper through multi-turn natural-language dialogue, integrating complementary observations across both views to arrive at the correct answer (see~\cref{fig:teaserr}).

\textsc{Cosmic} comprises five tasks spanning \textsc{object-level}, \textsc{relation-level}, and \textsc{map-level} collaborative spatial skills. Importantly, this hierarchical decomposition serves as a diagnostic framework, isolating whether failures arise at the level of cross-view grounding, relational reasoning, or allocentric integration. Together, these levels enable fine-grained identification of bottlenecks in current MLLMs. The five tasks are organized across three levels as follows:

\noindent\textbf{\textsc{Object-Level.}} (i) \textsc{Anchor Recognition.} This task requires agents to identify which objects appear in both their views, testing the fundamental skill of establishing common reference objects across viewpoints. (ii) \textsc{Global Counting.} Building on anchor recognition, this task requires agents to aggregate object instances across views, correctly disambiguating which instances are shared across views and which are only visible to one agent to avoid double-counting or omissions (see~\cref{fig:benchmark} top for examples).

\noindent\textbf{\textsc{Relation-Level.}} (iii) \textsc{Relative Distance.} In this task, agents must infer which object is closest or farthest from a target object. The target and candidate objects are distributed across both agent's views, requiring agents to fuse their partial observations to reason about relative proximity. (iv) \textsc{Relative Direction.} In this task, 
the Answerer must infer the egocentric direction of a target object absent from its own view. Resolving this requires both agents to coordinate a cross-view perspective transformation from the Helper's allocentric descriptions into the Answerer's egocentric frame (\cref{fig:benchmark} middle).

\noindent\textbf{\textsc{Map-Level.}} (v) \textsc{Cognitive Mapping.} This task evaluates whether agents can communicate and combine their partial egocentric observations into a map-like allocentric representation of the environment (\cref{fig:benchmark} bottom). Specifically, the answerer is presented with a candidate top-down map and tasked with judging whether it accurately represents the spatial layout of the shared environment.

\noindent\textit{Task formulation.} All tasks in \textsc{Cosmic} are posed as multiple-choice questions with one correct answer and three carefully constructed distractors to discourage superficial heuristics (see the supplementary for details on distractor construction). The exception is \textsc{Cognitive Mapping}, which we frame as a binary judgment of whether a presented top-down map is correct or incorrect, rather than free-form map generation, since evaluating freely generated maps would require generative assessment beyond the reliable capabilities of current MLLMs.

\begin{figure*}[tb]
  \centering
  \includegraphics[width=\textwidth]{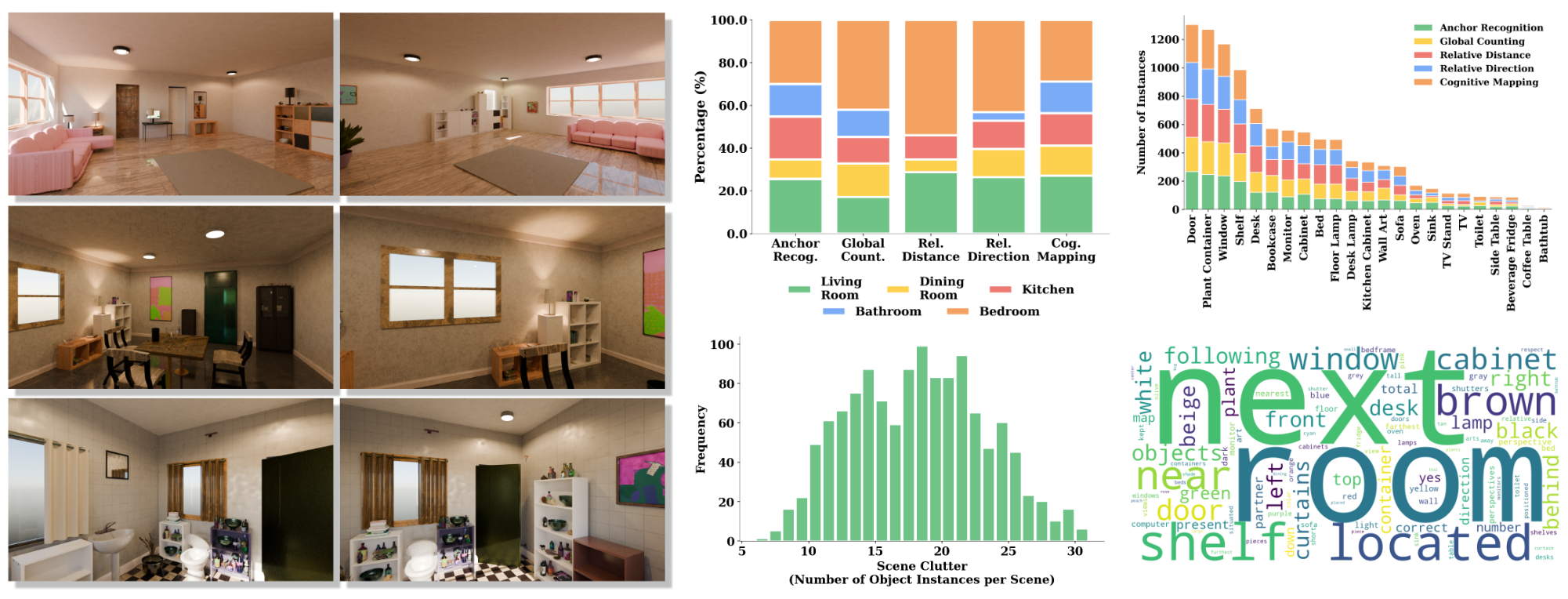}
  \caption{\textbf{\textsc{COSMIC} benchmark composition.} \textbf{Left:} Scenes from our benchmark. \textbf{Center Top:} Distribution of room types. \textbf{Center Bottom:} Distribution of scene clutter (number of object instances per scene). \textbf{Right Top:} Object-category frequencies across the benchmark. \textbf{Right Bottom:} Word cloud representing the most frequent spatial and object-related terms in the dataset.
    }
    \vspace{-4mm}
  \label{fig:data_stats}
\end{figure*}

\subsection{Benchmark Curation}

\noindent\textbf{Scene Generation.} To generate 3D indoor environments, we build upon Infinigen \cite{raistrick2024infinigen}, a procedural generation framework for synthesizing photorealistic 3D scenes. We extend it with a customized pipeline that enables fine-grained control over indoor spatial layouts, object distribution, and dual-view sampling. Viewpoints are sampled to ensure controlled partial overlap, with each pair of views sharing anchor objects while retaining objects exclusive to each perspective (\cref{fig:pipeline}). This produces high-fidelity scenes with complementary egocentric viewpoints tailored for collaborative reasoning, spanning diverse room types including Living Rooms, Bedrooms, Bathrooms, Kitchens, and Dining Rooms (\cref{fig:data_stats} left).

\noindent\textbf{Question Generation.} Given the object sets from both views, we first filter out visually ambiguous objects to prevent low-level perceptual noise from confounding spatial reasoning evaluation. Specifically, objects that are too small, indistinct, or partially occluded are excluded. Next, to uniquely refer to specific objects when constructing questions, each object is assigned descriptors based on its \textit{color}, \textit{size}, or \textit{neighboring objects}.  Descriptors are programmatically chosen to uniquely identify a single object instance within the scene, yielding references such as \textit{``purple door next to a yellow cabinet''}, \textit{``wall art above the cabinet''}, etc (\cref{fig:pipeline}). Objects that cannot be assigned a unique descriptor are excluded from question generation.
Finally, we employ fixed templates to generate questions spanning objects across both views ensuring that correct answers strictly require integrating information from both viewpoints. For instance, a global counting question \emph{``What is the total number of shelves in the room?''} is generated from a template \textit{``What is the total number of \texttt{<obj>} in the room?''}. The template-generated questions then undergo a paraphrasing stage to introduce linguistic diversity across the benchmark (see the supplementary for more details).

\noindent\textbf{Data Filtering and Human Verification.} To ensure that \textsc{Cosmic} serves as a rigorous evaluation of collaborative spatial reasoning, we implement a multi-stage filtering and verification pipeline. (i) \textsc{Cross-View Necessity Filtering.} We remove questions that can be answered through commonsense biases rather than genuine cross-view integration. For instance, an MLLM may correctly infer the relative direction of a center table by exploiting the prior that a sofa typically faces a center table, without requiring any information from the Helper's view. To filter out such cases, we prompt a strong MLLM ({\small Qwen3-VL-235B-A22B}) with only the Answerer's view for three trials per question, excluding any question the model answers correctly in all three runs. (ii) \textsc{Human Quality Assurance.} Following automated filtering, all remaining question-answer pairs undergo manual verification by the authors to ensure linguistic clarity, resolve spatial and color ambiguities, and confirm the correctness of ground-truth answers.



\noindent\textbf{Dataset Statistics.} \textsc{Cosmic} consists of 899 indoor room scenes paired with 1,250 unique question-answer instances, with 250 multiple-choice questions per task, each drawn from a unique scene. \cref{fig:data_stats} (center-top) shows the distribution of room types across tasks, reflecting sufficient scene diversity per subtask. The benchmark spans more than 23 distinct object categories, with Doors and Plant Containers being the most frequent (\cref{fig:data_stats} right-top). Scenes vary considerably in clutter, with a mean of 17.71 object instances per scene and ranging from 6 to 31 (\cref{fig:data_stats} center-bottom), adding to the scene-level complexity of the benchmark. Finally, the word cloud in \cref{fig:data_stats} (right-bottom) shows that spatial and object-related terms dominate the question vocabulary, reflecting the focus of our benchmark.

\begin{figure*}[tb]
  \centering

  \begin{minipage}{2\columnwidth}
    \raggedright
    \includegraphics[width=15.2cm,height=4.2cm]{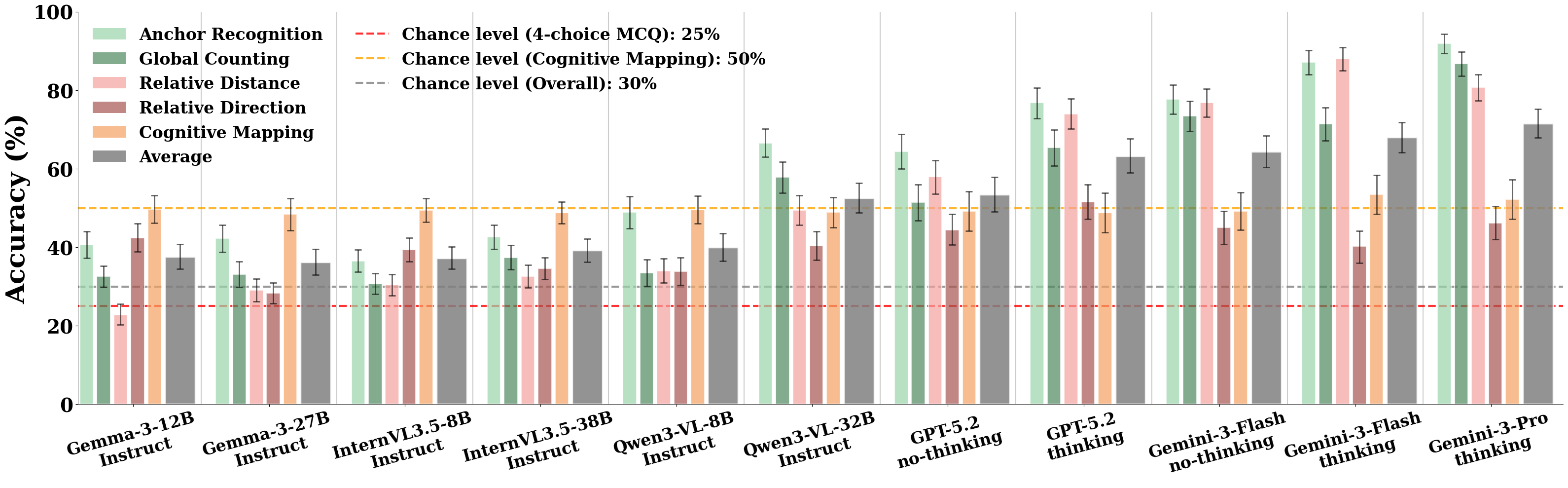}
  \end{minipage} 
  \\[1mm]
  \begin{minipage}{2\columnwidth}
    \raggedright
    \includegraphics[width=16.5cm,height=4.2cm]{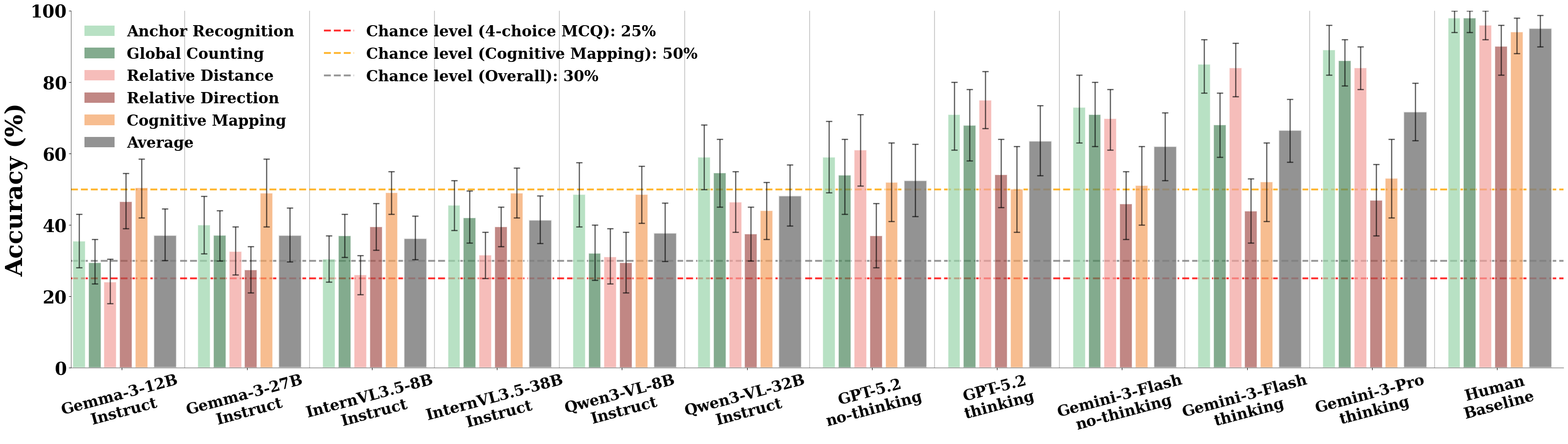}
  \end{minipage}
 
    \caption{
    \textbf{Top: Evaluation on \textsc{COSMIC}.} Error bars denote 90\% confidence intervals computed via bootstrap resampling. Dashed lines indicate chance levels (25\% for 4-choice MCQ, 50\% for binary map tasks and 30\% for overall).
    \textbf{Bottom: Evaluation on \textsc{COSMIC-Human}.}
    }
    \vspace{-4mm}
  \label{fig:main_result}
\end{figure*}

\section{Evaluation on the \textbf{\textbf{\textsc{Cosmic}}} Benchmark}

\noindent\textbf{Baseline Models.} We evaluate a range of recent state-of-the-art MLLM models. For open-source models, we include InternVL3.5 \cite{wang2025internvl3}, Qwen3-VL \cite{bai2025qwen3}, and Gemma-3 \cite{gemmateam2025gemma3technicalreport}. 
Among closed-source baselines we include GPT-5.2 \cite{singh2025openai}, Gemini-3-Flash and Gemini-3-Pro \cite{comanici2025gemini25pushingfrontier}. For  GPT-5.2 and Gemini-3-Flash, we report both \emph{no-thinking} and \emph{thinking} settings. 

\noindent\textbf{Evaluation Setup and Metric.} We conduct zero-shot evaluation of all models. Performance is measured by binary accuracy, where a response is considered correct if and only if the model selects the correct answer option. This applies uniformly across all tasks, including \textsc{Cognitive Mapping}. We report per-task accuracy with $90\%$ bootstrap confidence intervals, averaged over four runs for open-source models and two runs for closed-source models.

\noindent\textbf{Multi-turn Dialogue Protocol.} 
Each instance unfolds as a multi-turn conversation, initiated by the Answerer. The Helper responds based on its own view and the dialogue history. This alternating exchange continues until the Answerer decides it has gathered sufficient information and explicitly terminates the conversation, or until a fixed turn limit of 10 rounds is reached. Both agents have full access to the dialogue history throughout, and all communication proceeds exclusively through natural language with no parameter sharing or hidden state exchange between agents. Once the dialogue concludes, the Answerer is prompted to produce a final answer. For thinking models, we allow models to perform explicit intermediate reasoning before generating each dialogue message.

\noindent\textbf{Model Inputs and Role Conditioning.} Both Answerer and Helper agents are instantiated from the same underlying MLLM. Answerer agent's input includes \textsc{\{Image} $I_{A}$, \textsc{question} $q$, \textsc{answer options}, \textsc{task instruction}, \textsc{dialogue history\}}, with the additional inclusion of the candidate top-down map for \textsc{Cognitive Mapping} task. The Helper's input consists of \textsc{\{Image} $I_{H}$, \textsc{task instruction}, \textsc{dialogue history\}}. Role-conditioning prompts are appended to the system prompt (\textit{“You are the Answerer”, “You are the Helper”}) to enforce behavior specialization (see supp. for system prompts).

\noindent\textbf{Human Study.} 
We additionally collect \emph{human--human} dialogues under the same two-agent protocol on a subset of \textsc{Cosmic}, comprising 250 questions (50 per task), which we refer to as \textsc{Cosmic-Human}. This serves to establish (i) a human performance baseline and (ii) a basis for systematic comparison of model and human communication patterns. We conduct in-lab sessions with university students to gather these responses.
The interface mirrors the \textsc{Cosmic} setup, with Answerer and Helper roles, each observing only their egocentric view and jointly solving the task via multi-turn chat, subject to the same 10-round turn limit. Early termination is allowed. 
\section{Results on \textsc{Cosmic}}

\subsection{Main Results}

\noindent\textbf{MLLMs significantly underperform humans.}
\cref{fig:main_result} (Bottom) reports per-task accuracy on \textsc{Cosmic-Human}. Even the strongest model, Gemini-3-Pro-Think\-ing, achieves an average accuracy of only $71.82\%$, falling far short of the human baseline of $95.22\%$. This gap of over $23\%$ underscores that \textsc{Cosmic} poses a substantial challenge for current MLLMs, highlighting that \emph{communicating about space} remains a fundamentally difficult capability with significant room for improvement. Crucially, the gap is not uniform, it is narrowest on object-level tasks, where MLLMs show partial competence, but widens considerably on \textsc{Relative Direction} and \textsc{Cognitive Mapping}, where humans maintain near-ceiling accuracy. The contrast is most striking on \textsc{Cognitive Mapping}, where humans achieve $94\%$ accuracy while even frontier models perform near chance, suggesting that humans possess a form of collaborative spatial intelligence that current MLLMs fundamentally lack.

\noindent\textbf{Closed-source models consistently outperform open-source ones.}
Across all models on \textsc{Cosmic} (\cref{fig:main_result} Top), closed-source models consistently outperform open-source ones, with Gemini-3-Pro-Think\-ing and Gemini-3-Flash-Thinking ($71.64\%$, $67.88\%$) leading closed-source models, while Qwen3-VL-32B-Instruct (avg. $52.47\%$) leads open-source models, substantially outperforming InternVL\-3.5-38B (avg. $39.45\%$) and Gemma-3-27B (avg. $36.22\%$). Among the open-source models, scaling yields mixed results, Gemma-3 (12B: $37.43\%$ vs. 27B: $36.22\%$) and InternVL3.5 (8B: $37.39\%$ vs. 38B: $39.45\%$) show no statistically significant improvement from their smaller to larger checkpoints, as evidenced by overlapping confidence intervals, while Qwen3-VL benefits meaningfully from scale (8B: $40.08\%$ vs. 32B: $52.47\%$) with non-overlapping confidence intervals.


\begin{figure} 
    \centering
    \begin{minipage}{0.7\columnwidth}
    \centering
    \includegraphics[width=0.99\linewidth,height=4.4cm]{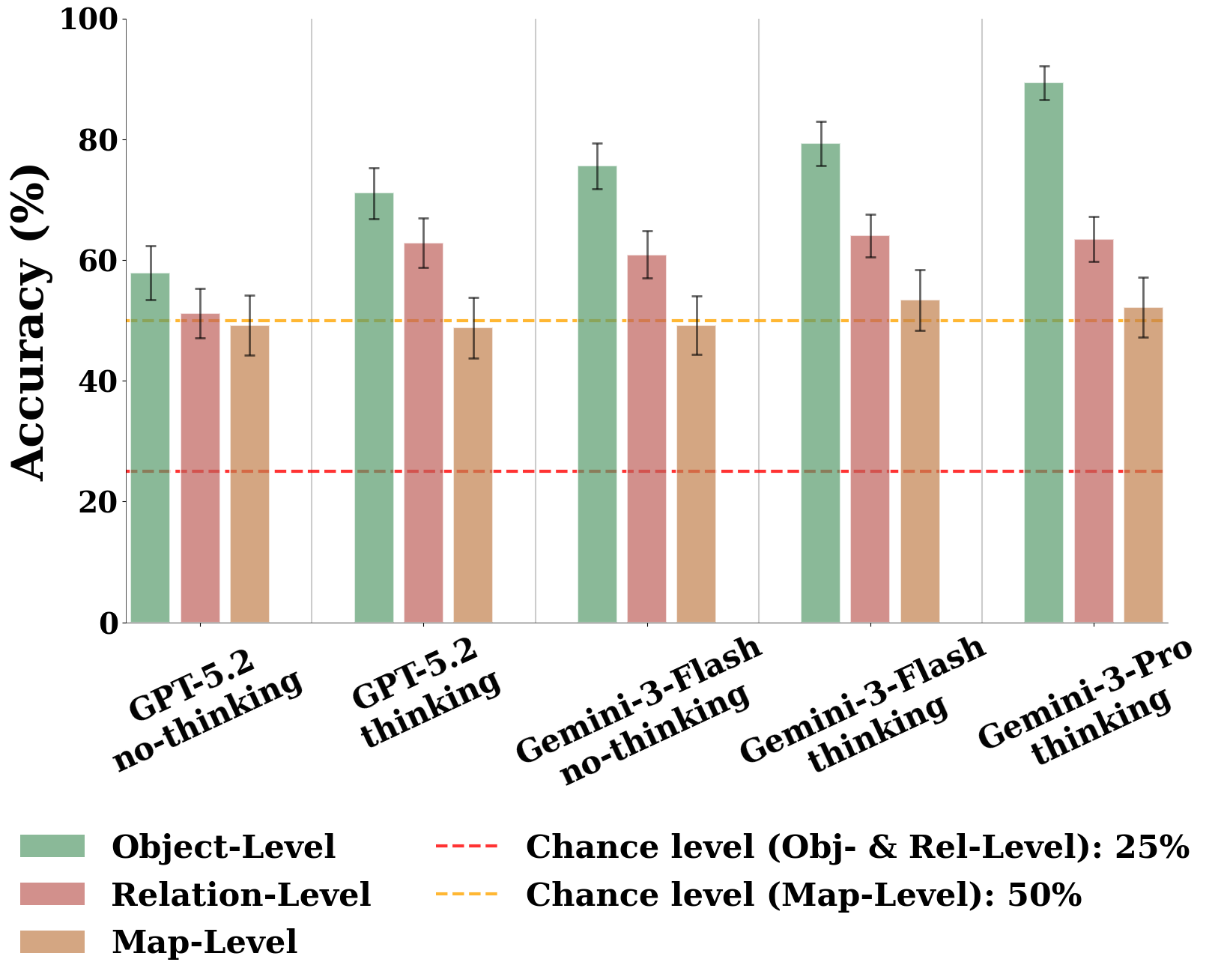}
      \end{minipage} 
      \begin{minipage}{0.28\columnwidth}
        \centering
        \includegraphics[width=0.99\linewidth,height=4.2cm]{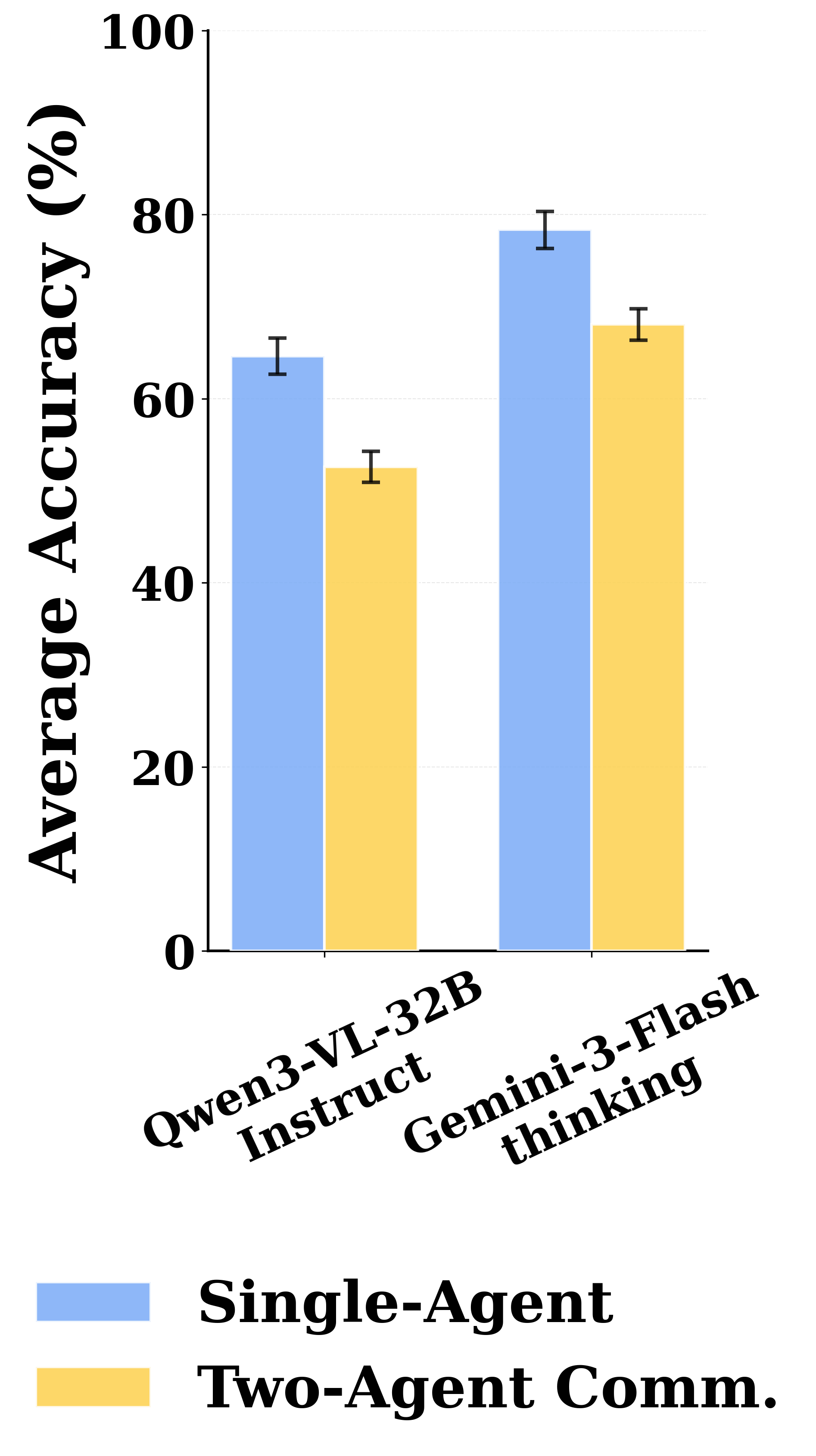}
      \end{minipage}
    \caption{\textbf{Left:} Performance Hierarchy in MLLMs across Object, Relation and Map-level. \textbf{Right:} Two-Agent Communication vs Single Agent.}
    \vspace{-15pt} 
    \label{fig:hier_perf}    
\end{figure}

\noindent\textbf{MLLMs' performance degrades from Anchors to Maps.}
For the closed-source systems, we observe a capability hierarchy across tasks, with model accuracy declining from object-level to map-level reasoning, with statistically significant gaps in most cases (\cref{fig:hier_perf} Left). We see in \cref{fig:main_result} (Top) that
\textsc{Anchor Recognition} is the easiest task for most models (Qwen3-VL-32B: $66.59\%$, GPT-5.2-Thinking: $76.83\%$, Gemini-3-Pro-Thin\-king: $91.99\%$), yet performance on even this most fundamental skill remains far from robust across the model spectrum. Performance degrades further on \textsc{Global Counting} and \textsc{Relative Distance}, which additionally require aggregating multiple object instances across views and cross-view metric reasoning respectively. 

The decline is steeper on \textsc{Relative Direction} (Gemini-3-Pro-Thinking: $46.21\%$, GPT-5.2-Thinking: $51.61\%$). 
Unlike \textsc{Relative Distance}, where agents can reason about proximity by comparing distances relative to shared anchor objects, \textsc{Relative Direction} requires the Answerer to transform the Helper's allocentric spatial descriptions, typically expressed relative to shared anchors, into its own egocentric frame of reference. This allocentric-to-egocentric transformation proves fundamentally challenging for current MLLMs, even the strongest models. \textsc{Cognitive Mapping} demands an even harder operation, jointly integrating both egocentric views into a globally consistent allocentric map of the full environment. Models collapse entirely on this task, with even frontier models near the $50\%$ chance baseline, indicating that full egocentric-to-allocentric integration across views remains beyond the reach of current MLLMs.

\noindent\textbf{Thinking helps object and metric reasoning but not geometric integration.} 
The capability hierarchy above raises a natural question of whether explicit reasoning can recover some performance at higher levels. We test this by enabling \emph{thinking} for GPT-5.2 and Gemini-3-Flash (see~\cref{fig:main_result} Top). \emph{Thinking} yields consistent, statistically significant gains on \textsc{Anchor Recognition} (Gemini-3-Flash $77.78\%$ vs. $87.19\%$, GPT-5.2 $64.34\%$ vs. $76.83\%$) and \textsc{Relative Distance} (Gemini-3-Flash $76.81\%$ vs. $88.00\%$, GPT-5.2 $58.01\%$ vs. $74.01\%$), suggesting that deliberate reasoning helps models more carefully match object descriptions and estimate metric relations across views. However, thinking yields no gain on \textsc{Relative Direction} or \textsc{Cognitive Mapping} for either model family (\cref{fig:main_result} Top). This dissociation reveals that the bottleneck at higher levels is not a failure of reasoning but a fundamental deficit in the geometric understanding required to reconcile egocentric observations into a shared spatial model. For \textsc{Global Counting}, the effect of \emph{thinking} is not consistent across models.


\noindent\textbf{Communication introduces additional difficulty beyond single-agent reasoning.} To isolate the impact of communication in collaborative spatial reasoning, we compare single-agent and two-agent performance for Qwen3-VL-32B-Instruct and Gemini-3-Flash-Thinking as representatives of open-source and closed-source model families respectively (\cref{fig:hier_perf} Right). 
In the single-agent setting, each model is given both egocentric views simultaneously and asked to answer directly without dialogue. 
Both models perform substantially worse when required to communicate, with Qwen3-VL-32B-Instruct dropping from $64.62\%$ to $52.62\%$ and Gemini-3-Flash-Thinking dropping from $78.38\%$ to $68.07\%$. This consistent performance gap indicates that the challenge of \textsc{Cosmic} is not solely attributable to the spatial reasoning tasks themselves, but is meaningfully compounded by the demands of coordinating through natural language and maintaining a coherent shared spatial model across turns. \emph{We believe that the time is now ripe to hold our MLLMs to higher standards by testing not just their spatial reasoning capabilities but also their capability to communicate that reasoning in natural language.} We hope that \textsc{Cosmic} will serve as a new standard benchmark for evaluating this capability.

\subsection{Failure Mode Analysis}

\begin{figure*}[tb]
  \centering
  \includegraphics[width=0.9\textwidth]{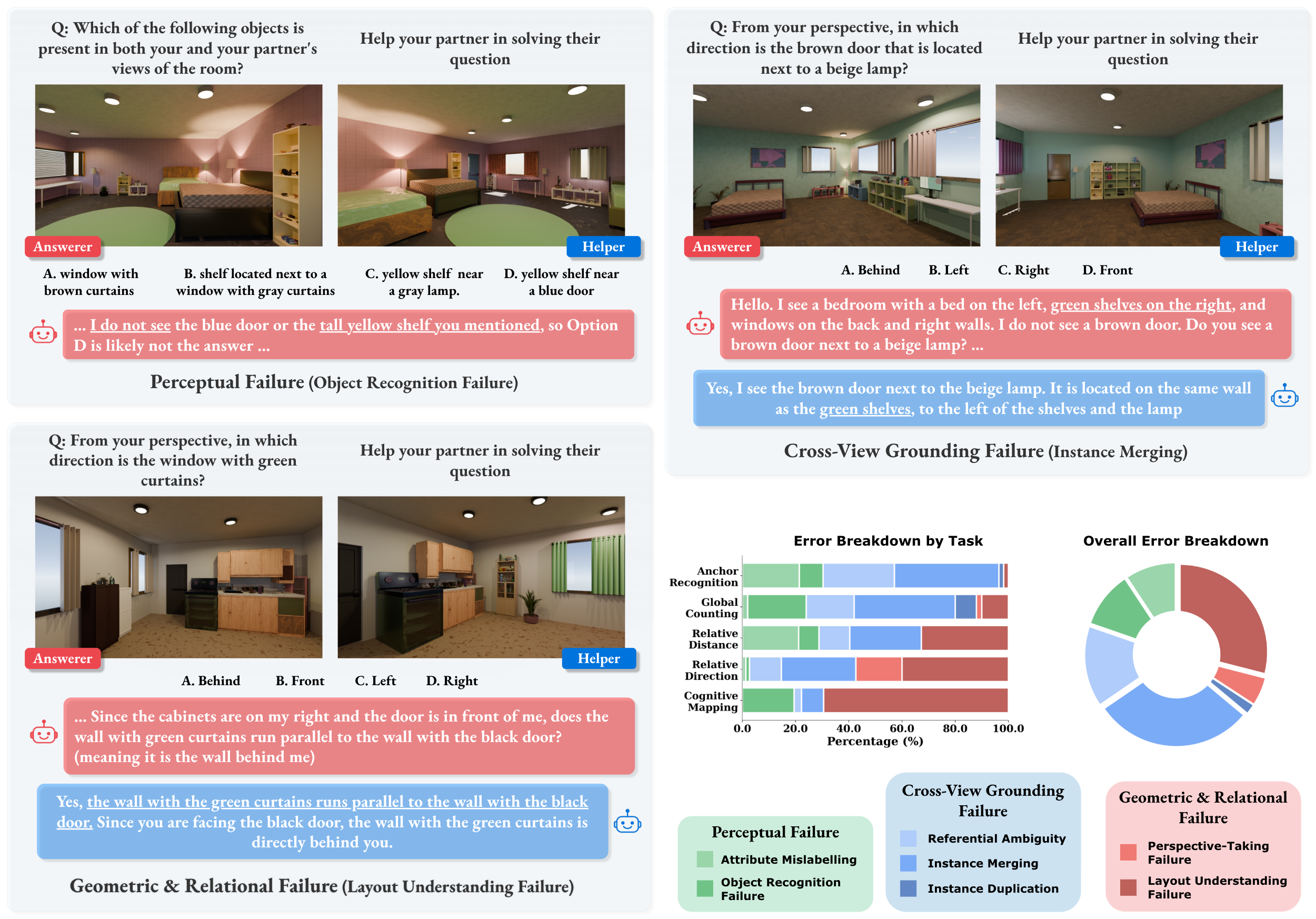}
  \caption{\textbf{Failure mode analysis on \textsc{COSMIC}.} Qualitative examples illustrate the three failure categories, Perceptual Failure (Object Recognition Failure), Cross-View Grounding Failure (Instance Merging), and Geometric and Relational Failure (Layout Understanding Failure). The bar chart shows the distribution of failure modes across tasks, and the donut chart shows the overall breakdown.} 
  \vspace{-4mm}
  \label{fig:error_analysis}
\end{figure*}

Why do frontier MLLMs fail at language-mediated spatial integration? To investigate, the authors manually examine agent--agent conversations and analyze where communication and reasoning breaks down. We review 150 failed instances (30 per task) for the best-performing model (Gemini-3-Pro-thinking), labeling each conversation with different error categories corresponding to failures at different stages of the dialogue (single conversation may exhibit multiple error types). We then compute the distribution of each error category over all errors made across the 150 conversations (\cref{fig:error_analysis}). We describe three main error categories below (see the supp. for more details).

\begin{enumerate}
    \item \textcolor{Green}{\textsc{Perceptual Failures.}} These failures encompass two subcategories. \textit{Object Recognition Failure} occurs when an agent entirely misses a visible object or misclassifies it as a different category, often triggered by environmental factors such as adverse lighting, cluttered backgrounds, or unusual viewing angles. \textit{Attribute Mislabelling} refers to an agent hallucinating or misidentifying object properties such as color or size, causing its partner to search for an object that does not match what is visible in their view (see~\cref{fig:error_analysis}). 
    \item \textcolor{Blue}{\textsc{Cross-View Grounding Failures.}} This category encompasses systemic breakdowns in establishing shared anchor objects across views. We identify three subcategories. \textit{Referential Ambiguity} arises when an agent generates utterances with underspecified object descriptions that fail to uniquely bind to a single instance in the partner's view, particularly in cluttered scenes. \textit{Instance Merging} occurs when two agents erroneously conclude that distinct object instances visible in their respective views refer to the same entity, collapsing them into one (see~\cref{fig:error_analysis} Top-Right). 
    \textit{Instance Duplication} is the converse failure, where a single entity visible in both views is treated as two separate instances, with each agent believing it refers to a different object.
    \item \textcolor{Red}{\textsc{Geometric \& Relational Failures.}} This category encompasses structural breakdowns in reasoning about spatial relations and transforming egocentric observations into a unified allocentric representation. \textit{Perspective-Taking Failures} occur when an agent incorrectly maps its partner's spatial descriptions onto its own egocentric frame of reference, resulting in systematic orientation errors such as left-right mirroring or front-back inversions. \textit{Layout Understanding Failure} occurs when an agent fails to reason about how objects are arranged relative to one another in 3D space from its 2D egocentric view, preventing it from constructing a coherent mental model of the scene's spatial layout (see~\cref{fig:error_analysis} Bottom-Left).
\end{enumerate}

\noindent\cref{fig:error_analysis} shows the distribution of failure modes across tasks. Perceptual Failures represent the smallest share of errors overall ($19.70\%$), suggesting that low-level scene perception is comparatively reliable. Cross-View Grounding Failures are the dominant error category overall ($46.09\%$), accounting for the largest share of failures, representing a primary bottleneck in collaborative spatial reasoning. They are most prominent on \textsc{Anchor Recognition} and \textsc{Global Counting} ($67.85\%$ \& $64\%$ respectively), consistent with the cross-view instance binding demands of these tasks.
Moving up the task hierarchy, Geometric and Relational Failures become increasingly dominant, constituting the vast majority of errors on \textsc{Relative Direction} and \textsc{Cognitive Mapping} ($57.33\%$ \& $69.44\%$ respectively), reflecting that these tasks evaluate a holistic understanding of the room's spatial layout that goes well beyond object-level reasoning. Note that cross-view grounding failures remain present across all tasks, consistent with it being a fundamental skill for any form of cross-view spatial reasoning. 

Overall, these results reveal a cascading failure dynamic where unresolved grounding errors can propagate through the dialogue and compound into geometric and relational failures that dominate on higher-level tasks. Since agents communicate exclusively through natural language, a single misidentified object or ambiguous reference early in the conversation can corrupt the shared spatial model that subsequent reasoning depends on.

\subsection{Human vs.\ Model Dialogue: How Different Is Collaboration?}
To investigate how the nature of communication differs between human-human and model-model, we compare the dialogues of human pairs on \textsc{Cosmic-Human} with those of MLLM agents.

\begin{figure}[tb]
  \centering
    \includegraphics[width=0.90\linewidth]{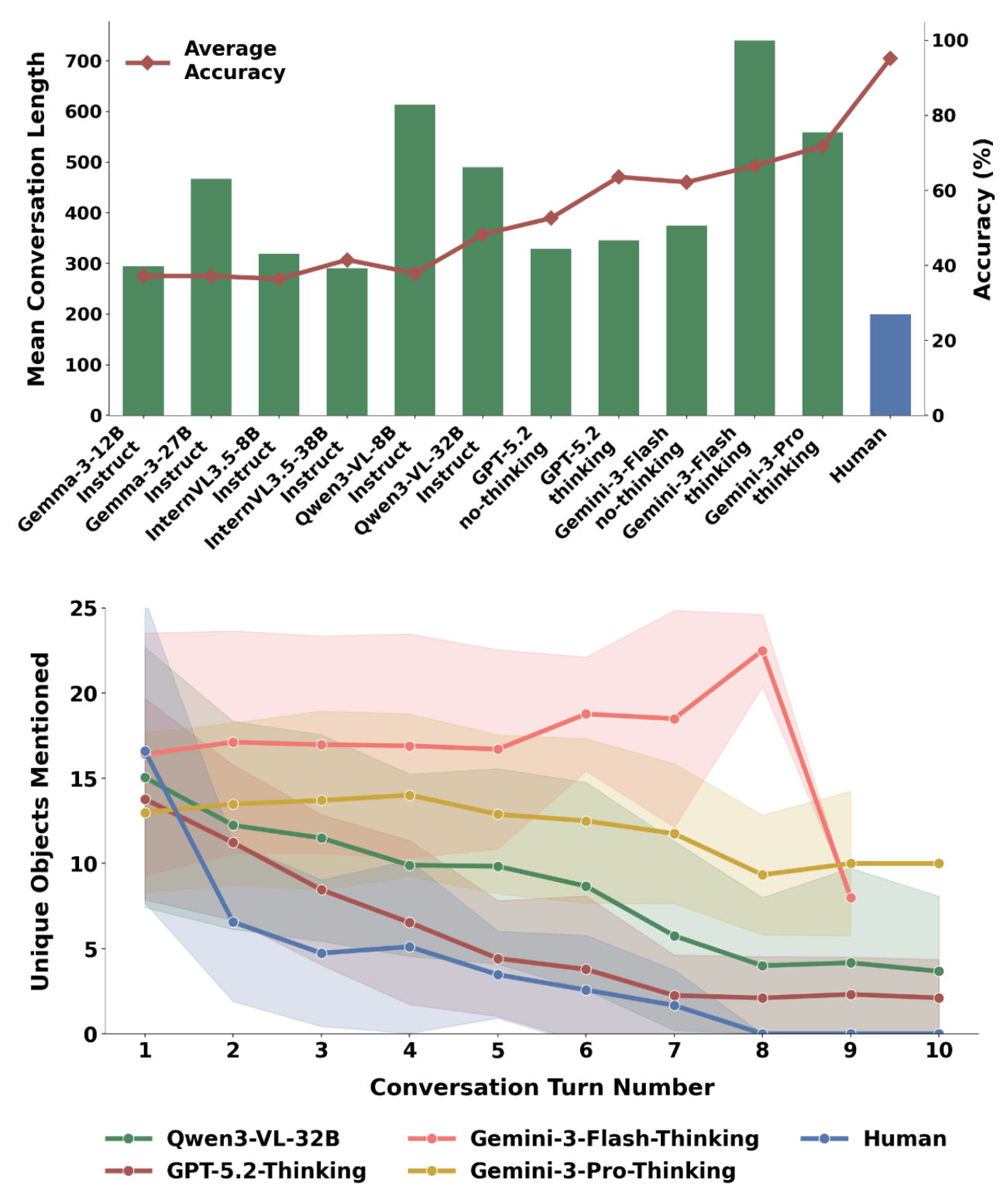}
    \caption{\textbf{Communication Efficiency \& Information Dynamics on \textsc{Cosmic-Human}.} \textbf{Top:} Communication Efficiency \textit{(Verbosity vs.\ Accuracy)}. \textbf{Bottom:} Information Dynamics \textit{(Unique Objects Mentioned vs. Conversation Turn Number)}. Shaded regions denotes variance across conversations.}  
    \vspace{-6mm}
  \label{fig:conv_analysis}
\end{figure}

\noindent\textbf{MLLMs Produce Verbose but Spatially Shallow Exchanges.} \cref{fig:conv_analysis} (Top) shows mean words per conversation against average accuracy for all models and human pairs. Notably, humans achieve the highest accuracy ($95.22\%$) while using least words per conversation (avg. $199.65$ words). In contrast, MLLM agents produce substantially more verbose exchanges (avg. across models $438.48$) yet achieve considerably lower accuracy (avg. across models $50.48\%$). Communication verbosity and accuracy are largely uncorrelated across models (pearson $r=0.37$, $p=0.26$), indicating no statistically significant relationship between the length of a dialogue and its effectiveness. For instance, Qwen3-VL-8B-Instruct generates the most words ($613.30$) among open-source models yet achieves among the lowest accuracies ($37.90\%$). This suggests that more verbose conversation does not necessarily lead to more effective conversation. We hypothesize that humans are efficient because of robust  spatial priors, built over a lifetime of navigating and communicating about physical environments, which support targeted, uncertainty-reducing exchanges.
In contrast, models compensate for weaker spatial representations with verbose but less informative dialogue.




\noindent\textbf{MLLMs Fail to Converge While Humans Systematically Narrow the Hypothesis Space.}
To understand how human and model dialogues differ in their information dynamics, we track the mean number of unique objects mentioned per turn across conversation turns (\cref{fig:conv_analysis} Bottom). Human pairs begin with a moderately high number of object mentions in the first turn and taper sharply over next turns, consistent with quickly locking onto a set of verified anchor objects and thereafter exchanging targeted, spatially precise updates to their mental model of the scene. This behavior is also qualitatively evident in \cref{fig:teaserr}, where human pairs quickly ground the sofa and peach window as shared anchors, and thereafter narrow down to targeted, spatially precise exchanges.
In contrast, MLLM agents show a slower and less consistent decline, with GPT-5.2-Thinking converging gradually and Gemini-3-Pro-Thinking and Gemini-3-Flash-Thinking sustaining a high rate of new object mentions throughout, as reflected in their persistently elevated curves in \cref{fig:conv_analysis} (Bottom). As also seen in \Cref{fig:teaserr}, rather than anchoring on shared objects early, MLLM agents persistently enumerate new scene elements (\textit{the white couch, black desk, white desk, and TV}), across every turn without converging on a shared reference frame from which to resolve the query. This pattern of continuous exploration rather than convergence directly explains the verbosity of MLLM conversations and their inability to build a coherent shared spatial model across turns.

\begin{figure} 
    \centering
    \vspace{-12pt} 
    \includegraphics[width=0.46\textwidth]{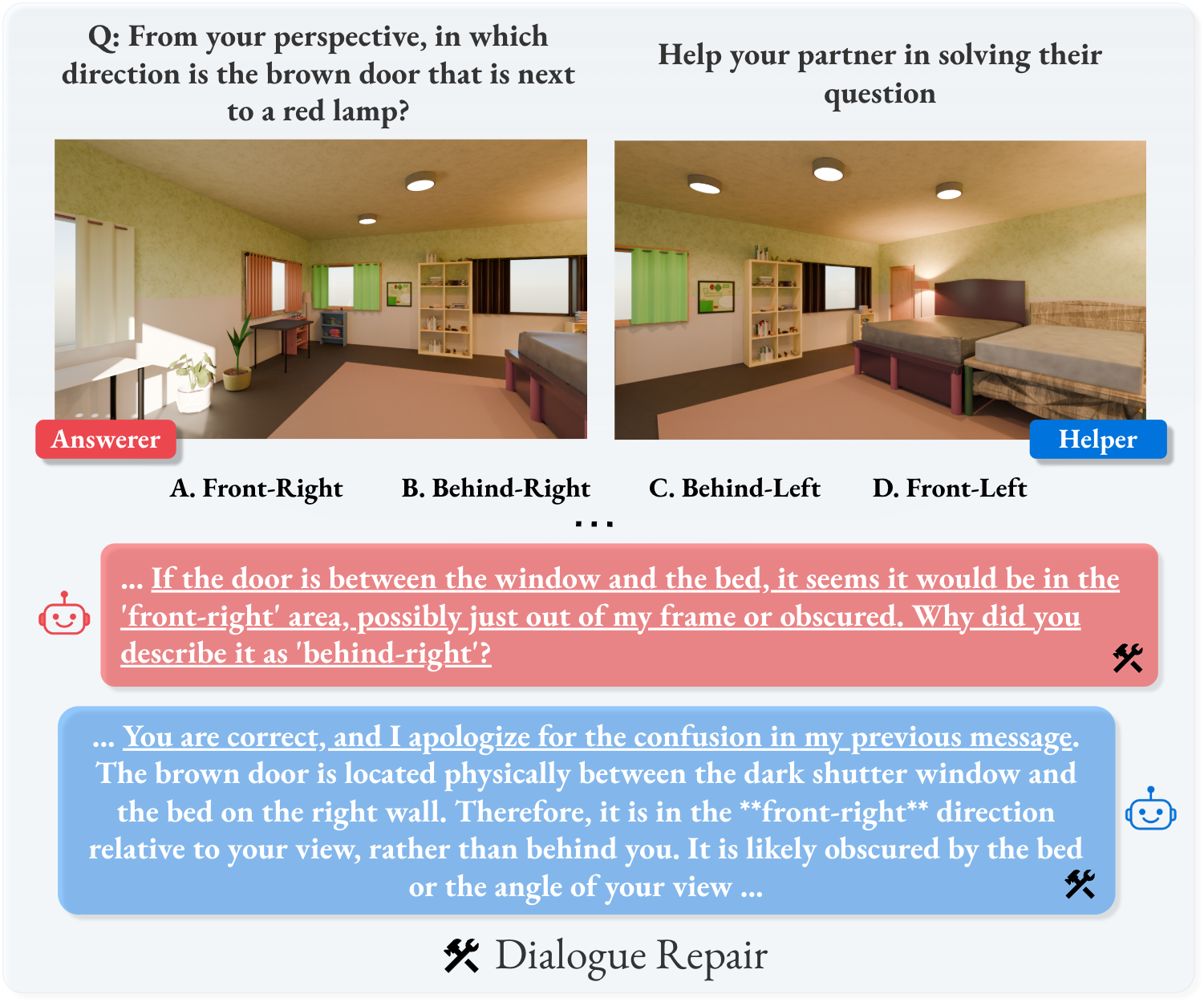}
    \caption{\textbf{Dialogue Repair behavior in Gemini-3-Pro-Thinking.}}
    \vspace{-4mm}
    \label{fig:dial_rep}
\end{figure}

\noindent\textbf{MLLMs Rarely Recover from Flawed Reasoning Trajectories.} 
Dialogue Repair refers to the metacognitive ability of agents to identify and correct erroneous reasoning trajectory during the conversation, a capacity that is critical for robustly building a shared spatial model.
We quantify such behavior in MLLMs and compare it against human pairs. For each conversation in \textsc{Cosmic-Human}, we employ a strong MLLM ({\small Gemini-3-Flash-Thinking}) as an automated judge, providing it with full task context including both egocentric views, the question \& options, the ground-truth answer, and the dialogue transcript (see supp). The judge produces a binary label indicating whether a repair event occurred (see~\cref{fig:dial_rep}). 

We define the dialogue repair rate as the fraction of conversations where agents successfully recovered from a wrong reasoning trajectory, out of all conversations where a wrong reasoning trajectory was observed (see supp.). Models exhibit far less dialogue repair than humans. Human pairs achieve a dialogue repair rate of $79.31\%$, while MLLM agents fall substantially short of this, with the best-performing model Gemini-3-Pro-Thinking reaching only $28.04\%$ and Qwen3-VL-32B achieving just $7.8\%$. 

Together, these analyses reveal that the failures of MLLMs reflect both a communication deficit and a deeper spatial reasoning deficit, with each compounding the other. Models struggle both to convey spatial observations effectively and to update their spatial model from their partner's utterances, a bidirectional breakdown that accumulates into an increasingly incoherent shared spatial representation.

\section{Conclusion and Future Work}
We introduced \textsc{Cosmic} and evaluated a broad set of frontier MLLMs. Our evaluation reveals a consistent capability hierarchy, current MLLMs show partial success at anchor grounding but deteriorate on relational reasoning and perform near chance on cognitive mapping. Human pairs achieve rapid convergence through targeted, information-dense exchanges, whereas MLLM agents resort to verbose dialogue that fails to yield a shared spatial model across turns.

\looseness=1
These findings point to several directions for future work. Progress on higher-level spatial tasks will likely require moving beyond linguistic chain-of-thought toward explicit visual reasoning mechanisms that support internal geometric verification and mental rotation. Structured spatial communication protocols such as anchor-first grounding conventions, explicit reference frame agreement, or sketch-based spatial descriptions, could help mitigate referential ambiguity and improve cross-view grounding. Finally, agents must develop proactive dialogue repair strategies to detect and correct flawed reasoning mid-conversation rather than accumulating contradictory spatial representations across turns. Addressing these gaps is essential for the development of capable, collaborative agents.
{
    \small
    \bibliographystyle{ieeenat_fullname}
    \bibliography{main}

@String(ICCV= {Int. Conf. Comput. Vis.})

@String(AAAI = {AAAI})

@String(ICCV  = {ICCV})

@article{CognitiveMaps1,
  title={Spatial cognition},
  author={Newcombe, Nora S},
  journal={Memory and Cognitive Processes},
  volume={3},
  pages={113--163},
  year={2004},
  publisher={Stevens’ Handbook of Experimental Psychology, John Wiley, New York, ed}
}

@article{garrod1987saying,
  title={Saying what you mean in dialogue: A study in conceptual and semantic co-ordination},
  author={Garrod, Simon and Anderson, Anthony},
  journal={Cognition},
  volume={27},
  number={2},
  pages={181--218},
  year={1987},
  publisher={Elsevier}
}

@article{badola2025MultiTurnPuzzles,
  title={Multi-turn puzzles: Evaluating interactive reasoning and strategic dialogue in llms},
  author={Badola, Kartikeya and Simon, Jonathan and Hosseini, Arian and Carthy, Sara Marie Mc and Munkhdalai, Tsendsuren and Goyal, Abhimanyu and Ko{\v{c}}isk{\`y}, Tom{\'a}{\v{s}} and Upadhyay, Shyam and Fatemi, Bahare and Kazemi, Mehran},
  journal={arXiv preprint arXiv:2508.10142},
  year={2025}
}

@inproceedings{tellex2011understanding,
  title={Understanding natural language commands for robotic navigation and mobile manipulation},
  author={Tellex, Stefanie and Kollar, Thomas and Dickerson, Steven and Walter, Matthew and Banerjee, Ashis and Teller, Seth and Roy, Nicholas},
  booktitle={Proceedings of the AAAI conference on artificial intelligence},
  volume={25},
  number={1},
  pages={1507--1514},
  year={2011}
}

@inproceedings{park2023generative,
  title={Generative agents: Interactive simulacra of human behavior},
  author={Park, Joon Sung and O'Brien, Joseph and Cai, Carrie Jun and Morris, Meredith Ringel and Liang, Percy and Bernstein, Michael S},
  booktitle={Proceedings of the 36th annual acm symposium on user interface software and technology},
  pages={1--22},
  year={2023}
}

@article{Tversky2003SpatialSchemas,
  title        = {Structures of Mental Spaces: How People Think About Space},
  author       = {Tversky, Barbara},
  journal      = {Environment and Behavior},
  volume       = {35},
  number       = {1},
  pages        = {66--80},
  year         = {2003}
}

@article{Levinson2003Space,
  title        = {Space in Language and Cognition: Explorations in Cognitive Diversity},
  author       = {Levinson, Stephen C.},
  journal      = {Language},
  volume       = {79},
  number       = {3},
  pages        = {620--622},
  year         = {2003}
}

@article{Clark1996UsingLanguage,
  title        = {Using Language},
  author       = {Clark, Herbert H.},
  journal      = {Cambridge University Press},
  year         = {1996}
}

@inproceedings{ThinkingInSpace,
  title={Thinking in space: How multimodal large language models see, remember, and recall spaces},
  author={Yang, Jihan and Yang, Shusheng and Gupta, Anjali W and Han, Rilyn and Fei-Fei, Li and Xie, Saining},
  booktitle={Proceedings of the Computer Vision and Pattern Recognition Conference},
  pages={10632--10643},
  year={2025}
}

@inproceedings{mincube,
  title={Spatial mental modeling from limited views},
  author={Yin, Baiqiao and Wang, Qineng and Zhang, Pingyue and Zhang, Jianshu and Wang, Kangrui and Wang, Zihan and Zhang, Jieyu and Chandrasegaran, Keshigeyan and Liu, Han and Krishna, Ranjay and others},
  booktitle={Structural Priors for Vision Workshop at ICCV'25},
  year={2025}
}

@article{mmsi,
  title={Mmsi-bench: A benchmark for multi-image spatial intelligence},
  author={Yang, Sihan and Xu, Runsen and Xie, Yiman and Yang, Sizhe and Li, Mo and Lin, Jingli and Zhu, Chenming and Chen, Xiaochen and Duan, Haodong and Yue, Xiangyu and others},
  journal={arXiv preprint arXiv:2505.23764},
  year={2025}
}

@article{seeingfromanother,
  title={Seeing from another perspective: Evaluating multi-view understanding in mllms},
  author={Yeh, Chun-Hsiao and Wang, Chenyu and Tong, Shengbang and Cheng, Ta-Ying and Wang, Ruoyu and Chu, Tianzhe and Zhai, Yuexiang and Chen, Yubei and Gao, Shenghua and Ma, Yi},
  journal={arXiv preprint arXiv:2504.15280},
  year={2025}
}

@article{tolman1948cognitivemaps,
  title={Cognitive maps in rats and men.},
  author={Tolman, Edward C},
  journal={Psychological review},
  volume={55},
  number={4},
  pages={189},
  year={1948},
  publisher={American Psychological Association}
}

@book{o1978hippocampus,
  title={The hippocampus as a cognitive map},
  author={O'keefe, John and Nadel, Lynn},
  year={1978},
  publisher={Oxford university press}
}

@article{viewspatial,
  title={Viewspatial-bench: Evaluating multi-perspective spatial localization in vision-language models},
  author={Li, Dingming and Li, Hongxing and Wang, Zixuan and Yan, Yuchen and Zhang, Hang and Chen, Siqi and Hou, Guiyang and Jiang, Shengpei and Zhang, Wenqi and Shen, Yongliang and others},
  journal={arXiv preprint arXiv:2505.21500},
  year={2025}
}

@article{li2025viewspatial,
  title={Viewspatial-bench: Evaluating multi-perspective spatial localization in vision-language models},
  author={Li, Dingming and Li, Hongxing and Wang, Zixuan and Yan, Yuchen and Zhang, Hang and Chen, Siqi and Hou, Guiyang and Jiang, Shengpei and Zhang, Wenqi and Shen, Yongliang and others},
  journal={arXiv preprint arXiv:2505.21500},
  year={2025}
}

@article{lee2025spatialmosaic,
  title={SpatialMosaic: A Multiview VLM Dataset for Partial Visibility},
  author={Lee, Kanghee and Lee, Injae and Kwak, Minseok and Ryu, Kwonyoung and Hong, Jungi and Park, Jaesik},
  journal={arXiv preprint arXiv:2512.23365},
  year={2025}
}

@article{ossowski2024comma,
  title={Comma: A communicative multimodal multi-agent benchmark},
  author={Ossowski, Timothy and Maqbool, Danyal and Chen, Jixuan and Cai, Zefan and Bradshaw, Tyler and Hu, Junjie},
  journal={arXiv preprint arXiv:2410.07553},
  year={2024}
}

@inproceedings{raistrick2024infinigen,
  title={Infinigen indoors: Photorealistic indoor scenes using procedural generation},
  author={Raistrick, Alexander and Mei, Lingjie and Kayan, Karhan and Yan, David and Zuo, Yiming and Han, Beining and Wen, Hongyu and Parakh, Meenal and Alexandropoulos, Stamatis and Lipson, Lahav and others},
  booktitle={Proceedings of the IEEE/CVF Conference on Computer Vision and Pattern Recognition},
  pages={21783--21794},
  year={2024}
}

@article{bai2025qwen3,
  title={Qwen3-vl technical report},
  author={Bai, Shuai and Cai, Yuxuan and Chen, Ruizhe and Chen, Keqin and Chen, Xionghui and Cheng, Zesen and Deng, Lianghao and Ding, Wei and Gao, Chang and Ge, Chunjiang and others},
  journal={arXiv preprint arXiv:2511.21631},
  year={2025}
}

@article{wang2025internvl3,
  title={Internvl3. 5: Advancing open-source multimodal models in versatility, reasoning, and efficiency},
  author={Wang, Weiyun and Gao, Zhangwei and Gu, Lixin and Pu, Hengjun and Cui, Long and Wei, Xingguang and Liu, Zhaoyang and Jing, Linglin and Ye, Shenglong and Shao, Jie and others},
  journal={arXiv preprint arXiv:2508.18265},
  year={2025}
}

@article{singh2025openai,
  title={Openai gpt-5 system card},
  author={Singh, Aaditya and Fry, Adam and Perelman, Adam and Tart, Adam and Ganesh, Adi and El-Kishky, Ahmed and McLaughlin, Aidan and Low, Aiden and Ostrow, AJ and Ananthram, Akhila and others},
  journal={arXiv preprint arXiv:2601.03267},
  year={2025}
}

@article{gemmateam2025gemma3technicalreport,
    title={Gemma 3},
    url={https://arxiv.org/abs/2503.19786},
    publisher={Google DeepMind},
    author={Gemma Team},
    year={2025}
}

@article{comanici2025gemini25pushingfrontier,
  title={Gemini 2.5: Pushing the frontier with advanced reasoning, multimodality, long context, and next generation agentic capabilities},
  author={Comanici, Gheorghe and Bieber, Eric and Schaekermann, Mike and Pasupat, Ice and Sachdeva, Noveen and Dhillon, Inderjit and Blistein, Marcel and Ram, Ori and Zhang, Dan and Rosen, Evan and others},
  journal={arXiv preprint arXiv:2507.06261},
  year={2025}
}

@misc{liu2023visualspatialreasoning,
      title={Visual Spatial Reasoning}, 
      author={Fangyu Liu and Guy Emerson and Nigel Collier},
      year={2023},
      eprint={2205.00363},
      archivePrefix={arXiv},
      primaryClass={cs.CL},
      url={https://arxiv.org/abs/2205.00363}, 
}

@misc{stogiannidis2025mindgapbenchmarkingspatial,
      title={Mind the Gap: Benchmarking Spatial Reasoning in Vision-Language Models}, 
      author={Ilias Stogiannidis and Steven McDonagh and Sotirios A. Tsaftaris},
      year={2025},
      eprint={2503.19707},
      archivePrefix={arXiv},
      primaryClass={cs.CV},
      url={https://arxiv.org/abs/2503.19707}, 
}

@misc{gholami2025spatialreasoningvisionlanguagemodels,
      title={Spatial Reasoning with Vision-Language Models in Ego-Centric Multi-View Scenes}, 
      author={Mohsen Gholami and Ahmad Rezaei and Zhou Weimin and Sitong Mao and Shunbo Zhou and Yong Zhang and Mohammad Akbari},
      year={2025},
      eprint={2509.06266},
      archivePrefix={arXiv},
      primaryClass={cs.CV},
      url={https://arxiv.org/abs/2509.06266}, 
}

@misc{yin2025mindcube,
      title={Spatial Mental Modeling from Limited Views}, 
      author={Baiqiao Yin and Qineng Wang and Pingyue Zhang and Jianshu Zhang and Kangrui Wang and Zihan Wang and Jieyu Zhang and Keshigeyan Chandrasegaran and Han Liu and Ranjay Krishna and Saining Xie and Manling Li and Jiajun Wu and Li Fei-Fei},
      year={2025},
      eprint={2506.21458},
      archivePrefix={arXiv},
      primaryClass={cs.AI},
      url={https://arxiv.org/abs/2506.21458}, 
}

@misc{gao2026map2thoughtexplicit3dspatial,
      title={Map2Thought: Explicit 3D Spatial Reasoning via Metric Cognitive Maps}, 
      author={Xiangjun Gao and Zhensong Zhang and Dave Zhenyu Chen and Songcen Xu and Long Quan and Eduardo Pérez-Pellitero and Youngkyoon Jang},
      year={2026},
      eprint={2601.11442},
      archivePrefix={arXiv},
      primaryClass={cs.CV},
      url={https://arxiv.org/abs/2601.11442}, 
}

@inproceedings{chen2024reconcile,
  title={Reconcile: Round-table conference improves reasoning via consensus among diverse llms},
  author={Chen, Justin and Saha, Swarnadeep and Bansal, Mohit},
  booktitle={Proceedings of the 62nd Annual Meeting of the Association for Computational Linguistics (Volume 1: Long Papers)},
  pages={7066--7085},
  year={2024}
}

@inproceedings{chen2024comm,
  title={Comm: Collaborative multi-agent, multi-reasoning-path prompting for complex problem solving},
  author={Chen, Pei and Zhang, Shuai and Han, Boran},
  booktitle={Findings of the Association for Computational Linguistics: NAACL 2024},
  pages={1720--1738},
  year={2024}
}

@inproceedings{du2024MAD,
  title={Improving factuality and reasoning in language models through multiagent debate},
  author={Du, Yilun and Li, Shuang and Torralba, Antonio and Tenenbaum, Joshua B and Mordatch, Igor},
  booktitle={Forty-first international conference on machine learning},
  year={2024}
}

@article{li2025hiddenbench,
  title={HiddenBench: Assessing Collective Reasoning in Multi-Agent LLMs via Hidden Profile Tasks},
  author={Li, Yuxuan and Naito, Aoi and Shirado, Hirokazu},
  journal={arXiv preprint arXiv:2505.11556},
  year={2025}
}

@article{xu2025vsbench,
  title={VS-Bench: Evaluating VLMs for Strategic Reasoning and Decision-Making in Multi-Agent Environments},
  author={Xu, Zelai and Xu, Zhexuan and Yi, Xiangmin and Yuan, Huining and Chen, Xinlei and Wu, Yi and Yu, Chao and Wang, Yu},
  year={2025}
}

@inproceedings{zhang2025mmcnav,
  title={MMCNav: MLLM-empowered Multi-agent Collaboration for Outdoor Visual Language Navigation},
  author={Zhang, Ziheng and Chen, Minghao and Zhu, Suguo and Han, Tingting and Yu, Zhou},
  booktitle={Proceedings of the 2025 International Conference on Multimedia Retrieval},
  pages={1767--1776},
  year={2025}
}

@article{zhu2026cavln,
  title={CA-VLN: Collaborative Agents in MLLM-Powered Visual-Language Navigation},
  author={Zhu, Ruolin and Li, Shaobin and Zhu, Zixing and Jia, Jing and Yang, Min},
  journal={Sensors (Basel, Switzerland)},
  volume={26},
  number={4},
  pages={1254},
  year={2026}
}

@inproceedings{kamath2023whatsup,
  title={What’s “up” with vision-language models? investigating their struggle with spatial reasoning},
  author={Kamath, Amita and Hessel, Jack and Chang, Kai-Wei},
  booktitle={Proceedings of the 2023 Conference on Empirical Methods in Natural Language Processing},
  pages={9161--9175},
  year={2023}
}

@misc{cheng2024spatialrgptgroundedspatialreasoning,
      title={SpatialRGPT: Grounded Spatial Reasoning in Vision Language Models}, 
      author={An-Chieh Cheng and Hongxu Yin and Yang Fu and Qiushan Guo and Ruihan Yang and Jan Kautz and Xiaolong Wang and Sifei Liu},
      year={2024},
      eprint={2406.01584},
      archivePrefix={arXiv},
      primaryClass={cs.CV},
      url={https://arxiv.org/abs/2406.01584}, 
}

@misc{jia2026omnispatialcomprehensivespatialreasoning,
      title={OmniSpatial: Towards Comprehensive Spatial Reasoning Benchmark for Vision Language Models}, 
      author={Mengdi Jia and Zekun Qi and Shaochen Zhang and Wenyao Zhang and Xinqiang Yu and Jiawei He and He Wang and Li Yi},
      year={2026},
      eprint={2506.03135},
      archivePrefix={arXiv},
      primaryClass={cs.CV},
      url={https://arxiv.org/abs/2506.03135}, 
}
}
\clearpage
\setcounter{page}{1}
\maketitlesupplementary

\definecolor{prompt1}{RGB}{223, 223, 192}
\definecolor{prompt1-frame}{RGB}{137, 137, 90}
\definecolor{prompt2}{RGB}{180, 230, 210}
\definecolor{prompt2-frame}{RGB}{70, 140, 120}
\definecolor{prompt3}{RGB}{212, 238, 179}
\definecolor{prompt3-frame}{RGB}{117, 146, 77}

\tcbset{
    prompt1/.style n args={1}{
        enhanced,
        breakable,
        colback=prompt1,        
        colframe=prompt1-frame,           
        fontupper=\ttfamily,      
        boxrule=1pt,              
        arc=2mm,                  
        left=1mm, right=1mm, top=1mm, bottom=1mm, 
        boxsep=4pt,               
        before skip=10pt, after skip=10pt, 
        overlay unbroken and first={
            \node[
                anchor=north west,
                xshift=4pt,
                text=white
            ] at (frame.north west) {\faFileLines};
        },
        title={~~~~\textbf{#1}},
        coltitle=white,
        fonttitle=\bfseries\small
    }
}

\tcbset{
    prompt2/.style n args={1}{
        enhanced,
        breakable,
        colback=prompt2,        
        colframe=prompt2-frame, 
        fontupper=\ttfamily,    
        boxrule=1pt,            
        arc=2mm,                
        left=1mm, right=1mm, top=1mm, bottom=1mm,
        boxsep=4pt,
        before skip=10pt, after skip=10pt,
        overlay unbroken and first={
            \node[
                anchor=north west,
                xshift=4pt,
                text=white
            ] at (frame.north west) {\faFileLines};
        },
        title={~~~~\textbf{#1}},   
        coltitle=white,
        fonttitle=\bfseries\small
    }
}

\tcbset{
    prompt3/.style n args={1}{
        enhanced,
        breakable,
        colback=prompt3,        
        colframe=prompt3-frame,            
        fontupper=\ttfamily,               
        boxrule=1pt,                       
        arc=2mm,                           
        left=1mm, right=1mm, top=1mm, bottom=1mm, 
        boxsep=4pt,                        
        before skip=10pt, after skip=10pt, 
        overlay unbroken and first={
            \node[
                anchor=north west,
                xshift=4pt,
                text=white
            ] at (frame.north west) {\faFileLines};
        },
        title={~~~~\textbf{#1}},
        coltitle=white,
        fonttitle=\bfseries\small
    }
}

{\Large \textbf{Overview of Supplementary}}\\
\begin{enumerate}
    \item \textbf{\textsc{Cosmic} Benchmark Design} (\cref{supp:S1})
    \item \textbf{Evaluation Protocol and Reproducibility} (\cref{supp:S2})
    \item \textbf{Failure Mode Analysis} (\cref{supp:S3})
    \item \textbf{Human Data Collection Interface} (\cref{supp:S4})
    \item \textbf{Multi-turn Dialogue} (\cref{supp:S5})
    \item \textbf{Dialogue Repair} (\cref{supp:S6})
    \item \textbf{Model and Human Conversations} (\cref{supp:S7})
    \item \textbf{Case Study} (\cref{supp:S8})
    \item \textbf{Compute Resources} (\cref{supp:S9})
    \item \textbf{Broader Impact} (\cref{supp:S10})
    \item \textbf{Limitations} (\cref{supp:S11})
\end{enumerate}

\section{\textsc{Cosmic} Benchmark Design}
\label{supp:S1}
\begin{figure*}[t]
    \centering
    \includegraphics[width=\textwidth]{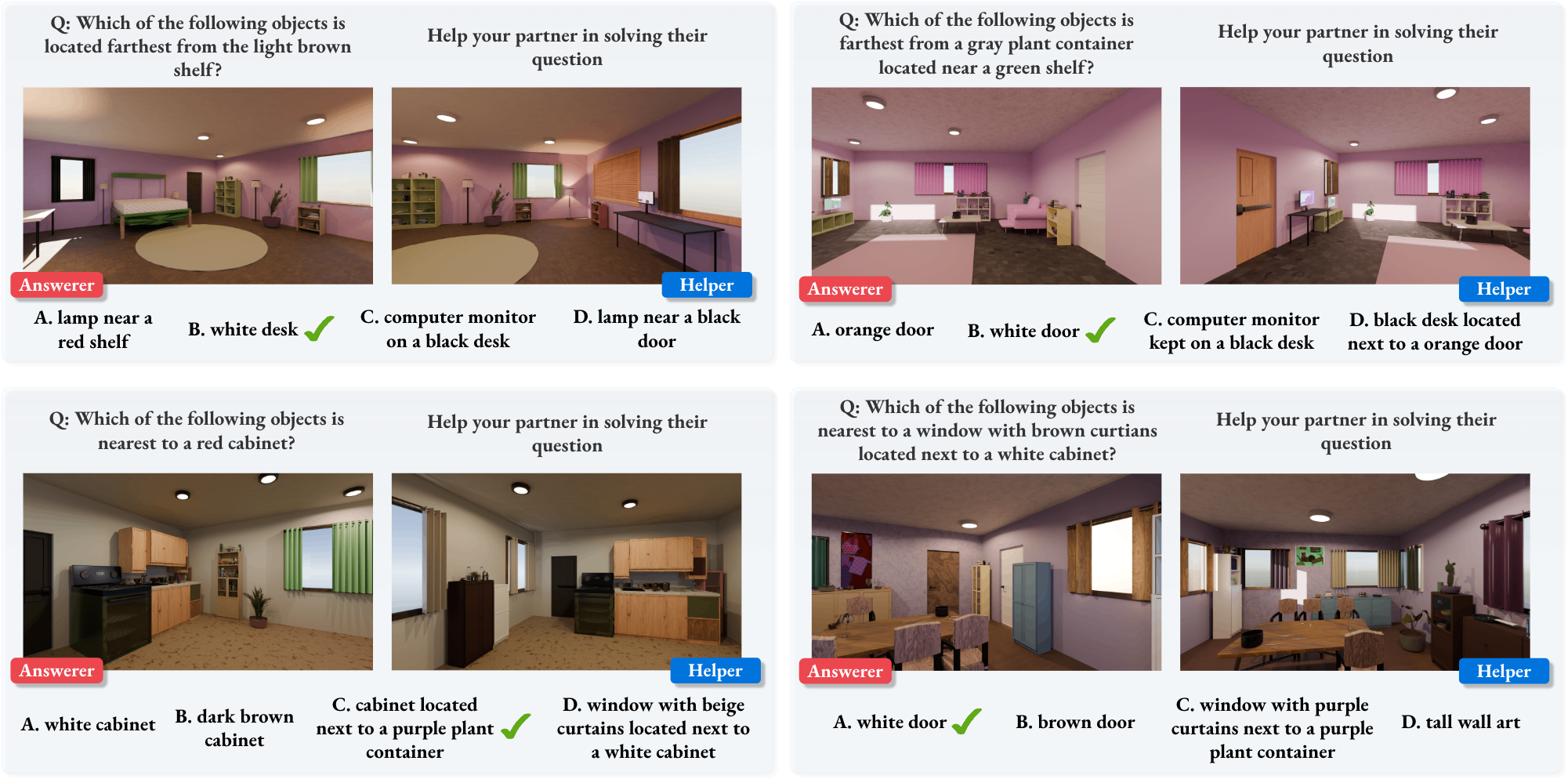}
    \caption{\textbf{Samples from \textsc{Cosmic} Benchmark (Relative Distance).}}
    \label{fig:dist_samples}
\end{figure*}

\begin{figure*}[t]
    \centering
    \includegraphics[width=\textwidth]{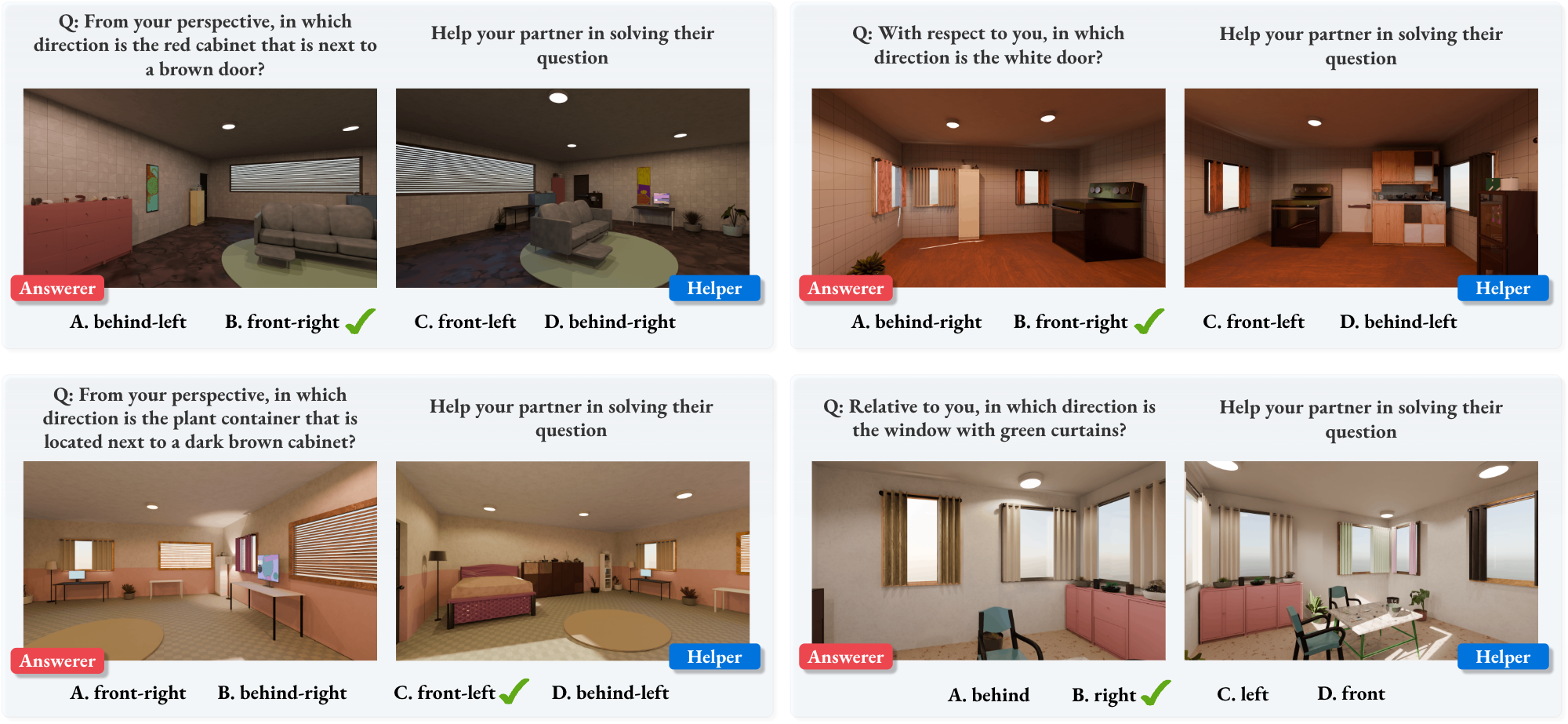}
    \caption{\textbf{Samples from \textsc{Cosmic} Benchmark (Relative Direction).}}
    \label{fig:direct_samples}
\end{figure*}

\begin{figure*}[t]
    \centering
    \includegraphics[width=\textwidth]{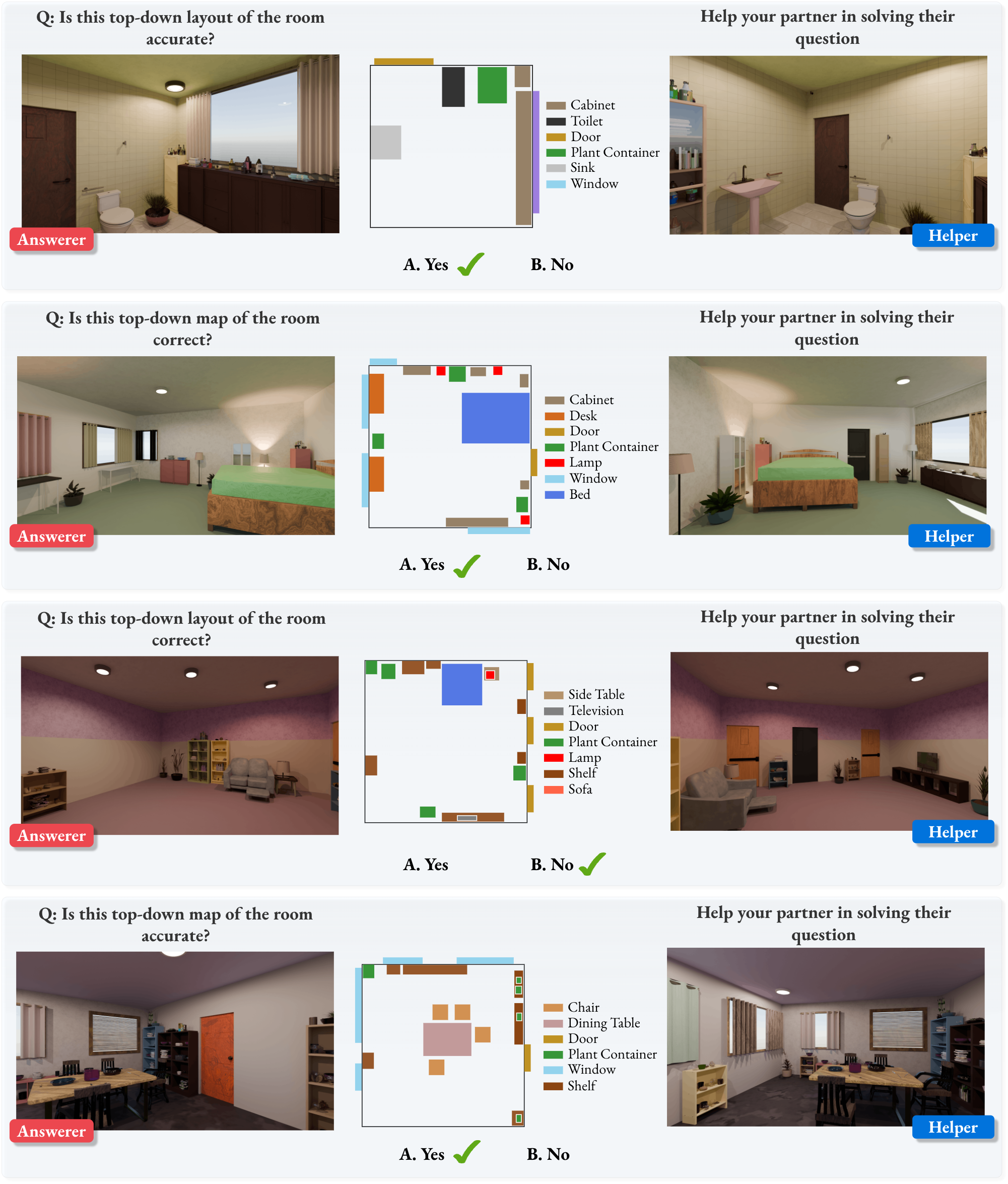}
    \caption{\textbf{Samples from \textsc{Cosmic} Benchmark (Cognitive Mapping).}}
    \label{fig:map_samples}
\end{figure*}

This section provides additional details on the \textsc{Cosmic} benchmark curation pipeline. We also provide several examples from our benchmark in~\cref{fig:anchor_samples,fig:counting_samples,fig:dist_samples,fig:direct_samples,fig:map_samples}.

\subsection{Scene Generation}
Each environment $\mathcal{E}$ contains a set of objects $O = {o_1, \dots, o_M}$, with the Answerer ($A$) and Helper ($H$) agents placed at distinct viewpoints. Answer and Helper receive their respective egocentric views $I_{A}$ and $I_{H}$. The set of visible objects in each view, $O_{A} = \{\, o_j \in O \mid o_j \text{ is visible in } I_{A} \,\}$ and $O_{H} = \{\, o_j \in O \mid o_j \text{ is visible in } I_{H} \,\}$, are designed such that their union ($O_{A} \cup O_{H} \approx O$) covers most of the environment, while their intersection ($O_{A} \cap O_{H} \neq \emptyset$) provides shared anchor objects that allows cross-view grounding and perspective alignment between agents. Overall, this formulation transforms spatial reasoning from a single-agent perception task into a \emph{collaborative spatial reasoning task.}

To ensure that the visibility sets $O_{A}$ and $O_{H}$ satisfy these properties by construction, we employ a controlled dual-view camera sampling strategy. Both cameras are placed at a fixed altitude of $1.5$m above the floor with a fixed horizontal pitch, producing natural egocentric views at human eye level. Yaw is sampled independently and uniformly over $[-180^{\circ}, 180^{\circ}]$ for each camera, ensuring diverse viewing directions across scenes. To determine $O_{A}$ and $O_{H}$, we employ a ray-casting visibility algorithm. For each mesh object in the scene, we project all eight corners of its axis-aligned bounding box into camera space and cast rays from the camera origin toward each corner. An object is considered visible from a given camera if at least three of its eight bounding box corners are both within the camera frustum and unoccluded by intervening geometry. Camera poses are resampled until the resulting visibility sets satisfy $O_{A} \cap O_{H} \neq \emptyset$ and $O_{A} \cup O_{H} \approx O$.

\subsection{Question Generation}
To generate questions for different tasks as described in~\cref{sec:qtypes}, we rely on predefined templates to produce the base question strings shown in~\cref{tab:tasks}. These template-generated questions are then passed through a LLM paraphrasing stage {\small gpt-4o-mini}, which rephrases them to introduce linguistic diversity and naturalness while preserving their semantic content.~\cref{tab:paraphrases} demonstrates examples of these paraphrases. 

\begin{table}[t]
\centering
\caption{Templates used for question generation. Below, \textit{\textbf{object category}} is a placeholder for an object category, for example \textit{sofa}, while \textit{\textbf{object description}} is a placeholder for the unique description of a given object instance, for example \textit{green sofa near a white door}.}
\label{tab:tasks}
\begin{tabular}{p{0.35\linewidth}@{\hspace{0.02\linewidth}}p{0.55\linewidth}}
\toprule
Task & Template \\
\midrule
Anchor Recognition & Which of the following objects is visible in both your and your partner's views of the room? \\
Global Counting & What is the total number of \textit{$<$object category$>$} in the room? \\
Relative Direction & From your perspective, in which direction is \textit{$<$object description$>$}? \\
Relative Distance & Which of the following objects is closest to / farthest from the \textit{$<$object description$>$}? \\
Cognitive Mapping & Is this top-down map of the room correct? \\
\bottomrule
\end{tabular}
\end{table}

\begin{table*}[t]
\centering
\caption{Examples of template-generated questions and their corresponding paraphrased versions.}
\label{tab:paraphrases}

\begin{tabular}
{p{0.45\linewidth}@{\hspace{0.04\linewidth}}p{0.45\linewidth}}
\toprule
\textbf{Before Paraphrasing} & \textbf{After Paraphrasing} \\
\midrule

\parbox[t]{\linewidth}{
\textbf{Question:} Which of the following objects is visible in both your and your partner's views of the room?\\
\textbf{Options:}\\
(A) floor purple Lamp\\
(B) black shutter Window\\
(C) green Plant Container next to a black Desk\\
(D) Lamp near a black shutter Window
}
&
\parbox[t]{\linewidth}{
\textbf{Question:} Which of the following objects is present in both your and your partner's views of the room?\\
\textbf{Options:}\\
(A) floor purple lamp\\
(B) window with black shutters\\
(C) green plant container located next to a black desk\\
(D) lamp near a window with black shutters
}
\\

\midrule

\parbox[t]{\linewidth}{
\textbf{Question:} What is the total number of Monitor in the room?\\
\textbf{Options:}\\
(A) 1\\
(B) 2\\
(C) 3\\
(D) 4
}
&
\parbox[t]{\linewidth}{
\textbf{Question:} What is the total number of computer monitors in the room?\\
\textbf{Options:}\\
(A) 1\\
(B) 2\\
(C) 3\\
(D) 4
}
\\

\midrule

\parbox[t]{\linewidth}{
\textbf{Question:} Which of the following objects is closest to the floor purple Lamp?\\
\textbf{Options:}\\
(A) beige curtain Window\\
(B) Monitor near a green Plant Container\\
(C) brown curtain Window\\
(D) green Plant Container next to a black Desk
}
&
\parbox[t]{\linewidth}{
\textbf{Question:} Which of the following objects is nearest to the floor lamp with purple color?\\
\textbf{Options:}\\
(A) window with beige curtains\\
(B) computer monitor located near a green plant container\\
(C) window with brown curtains\\
(D) green plant container located next to a black desk
}
\\

\midrule

\parbox[t]{\linewidth}{
\textbf{Question:} From your perspective, in which direction is Window near a Monitor?\\
\textbf{Options:}\\
(A) behind\\
(B) front\\
(C) left\\
(D) right
}
&
\parbox[t]{\linewidth}{
\textbf{Question:} From your perspective, in which direction is the window that is near a computer monitor?\\
\textbf{Options:}\\
(A) behind\\
(B) front\\
(C) left\\
(D) right
}
\\

\midrule

\parbox[t]{\linewidth}{
\textbf{Question:} Is this top-down map of the room correct?
}
&
\parbox[t]{\linewidth}{
\textbf{Question:} Is this top-down layout of the room accurate?
}
\\

\bottomrule
\end{tabular}
\end{table*}

\subsection{Distractor Design and Task Difficulty}
\label{sec:qtypes}

We describe the distractor design and the task difficulty for each task below:

\noindent\textbf{\textsc{Anchor Recognition.}}
The primary difficulty in this task lies in communicating the overlap between two egocentric observations precisely. An example question for this task is \textit{``Which of the following objects is present in both your and your partner's views of the room?''} The question template for this task is provided in~\cref{tab:tasks}.
Agents must dynamically reference objects via their attributes or relational cues (e.g., \textit{``the yellow cabinet next to the door''}), since multiple object instances of the same category with similar visual properties may appear across the two views, making category labels alone insufficient to determine whether two agents are referring to the same physical instance. This mirrors the referential grounding challenges that arise in real-world cluttered environments, where precise, discriminative descriptions are essential for establishing shared reference. 

\begin{figure*}[t]
    \centering
    \includegraphics[width=\textwidth]{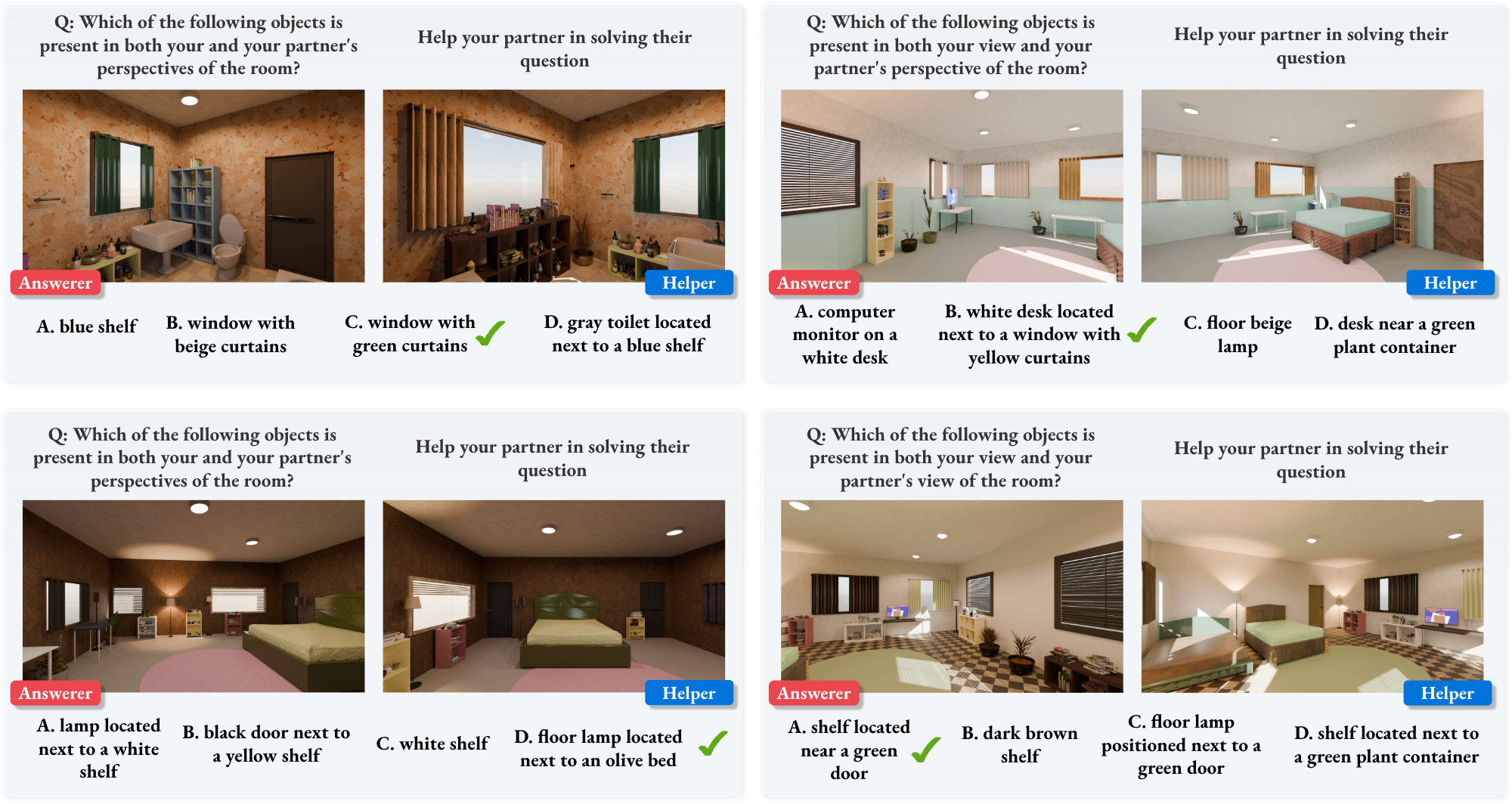}
    \caption{\textbf{Samples from \textsc{Cosmic} Benchmark (Anchor Recognition).}}
    \label{fig:anchor_samples}
\end{figure*}

The correct answer is the object present in $O_A \cap O_H$. We construct three distractors 
(see~\cref{fig:anchor_samples} top-left) as follows, (i) an object visible exclusively to the Answerer ($O_A \setminus O_H$) (e.g. \textit{window with green curtains}), (ii) an object visible exclusively to the Helper ($O_H \setminus O_A$) (e.g. \textit{blue shelf}), and (iii) an object from the same semantic 
category as the correct answer but differing in at least one discriminative attribute (e.g., color, size, or spatial relation). This third distractor, drawn exclusively from one agent's view, is specifically designed to prevent models from succeeding via category-level matching alone, requiring instead attribute- and relation-based reasoning (e.g. \textit{window with beige curtains}). When a scene does not contain a same-category object with differing attributes, this slot is filled by an object sampled from another available category in the scene (e.g. \textit{gray toilet located next to a blue shelf}). 

\noindent\textbf{\textsc{Global Counting.}}
The central difficulty in this task is \emph{cross-view deduplication}. The question template for this task is provided in~\cref{tab:tasks}, with a representative example being \textit{``What is the total number of shelves in the room?''}

Building on \textsc{Anchor Recognition}, in this task, agents must jointly determine which object instances across their two views refer to the same physical entity and which are distinct, communicating this through language to avoid both double-counting shared instances and omitting instances visible only to one agent. This demands simultaneous cross-view aggregation and instance-level disambiguation, requiring agents to produce and interpret precise, discriminative descriptions of objects within the same category.

\begin{figure*}[t]
    \centering
    \includegraphics[width=\textwidth]{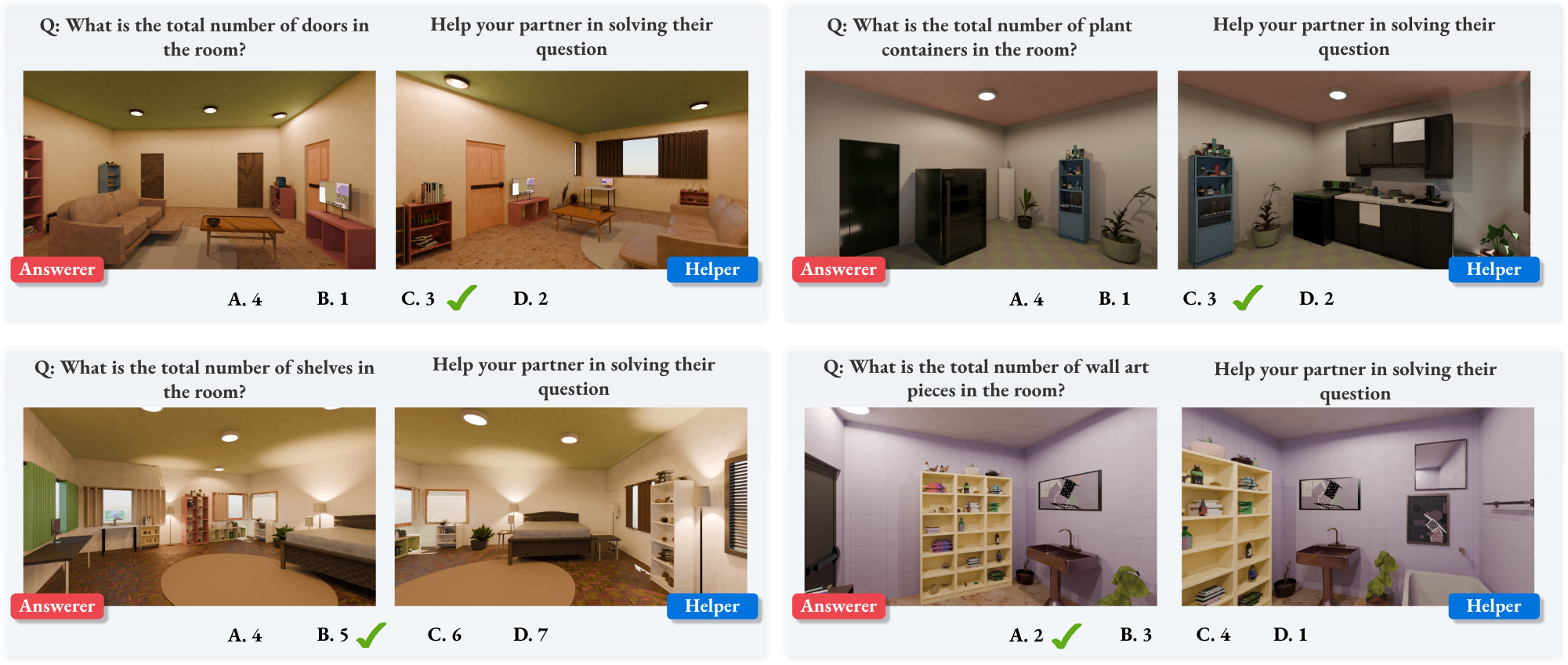}
    \caption{\textbf{Samples from \textsc{Cosmic} Benchmark (Global Counting).}}
    \label{fig:counting_samples}
\end{figure*}

Questions are generated by aggregating all instances of a target object category across both views and querying the total count, i.e., $|O_{\text{ans}} \cup O_{\text{help}}|$. Distractors are constructed to target two canonical failure modes (see~\cref{fig:counting_samples} top-left for an example), \emph{overcounting}, corresponding to the naive sum 
$|O_{\text{ans}}| + |O_{\text{help}}|$ (i.e., double-counting shared instances), and \emph{undercounting} (i.e., omitting certain instances). For remaining distractors, we sample a count value close to the ground truth.

\noindent\textbf{\textsc{Relative Distance.}}
The difficulty of this task stems from the need to compare metric distances across two spatially disjoint views. A typical question takes the form, \textit{``Which of the following objects is closest to the white computer desk?''} (see~\cref{tab:tasks} for question template). Agents cannot directly observe all candidate objects and must instead align their respective partial views through communication before a meaningful distance comparison is possible. 

To generate questions, we select an anchor object to be the \textit{target object}. We then sample four candidate objects for the options from each agent's exclusive view ($O_A \setminus O_H$ and $O_H \setminus O_A$). Distances are then computed between the bounding box of the \textit{target object} and each candidate. The correct answer is the closest (or farthest) candidate from the \textit{target object}. We enforce a minimum distance margin between the correct answer and the next closest candidate to ensure the difference is geometrically unambiguous.

\noindent\textbf{\textsc{Relative Direction.}}
This task requires a cross-view perspective transformation. An example question for this task is \textit{``From your perspective, in which direction is the computer desk located?''} where the \textit{target object} (computer desk) is visible only to the Helper. Answerer agent must infer the direction of the target object with respect to its own view by grounding via shared anchor objects and mapping its partner's spatial descriptions into its own frame of reference. The difficulty increases in scenes with little overlap across views, since the perspective transformation needs to be performed with limited shared grounding.


Answer options correspond to eight egocentric directions, \textit{Front, Front-Right, Right, Behind-Right, Behind, Behind-Left, Left,} and \textit{Front-Left}, representing angular offsets of $45^\circ$. A direction is assigned when the object's bearing falls within a $\pm 10^\circ$ window centered on the corresponding orientation angle. 

\noindent\textbf{\textsc{Cognitive Mapping.}}
This task requires agents to jointly construct a coherent allocentric (top-down) representation of the environment from their complementary egocentric views, making it the most demanding task in the benchmark. Rather than just reasoning about individual objects or pairwise relations, agents must integrate their partial observations into a globally consistent top-down map. A typical question takes the form, \textit{``Is this top-down map of the room correct?''} (see~\cref{tab:tasks} for the question template, subsequently paraphrased using an LLM). 

We construct the correct map by projecting object positions from the scene onto a 2D top-down grid. Distractor maps are generated by swapping the positions of Helper-exclusive objects in the correct map, producing layouts that contain the correct set of objects but placed at incorrect spatial locations. The options for this task are binary, i.e. either \textit{Yes} or \textit{No}. Maps are rendered as a visual image and provided directly to the Answerer agent as input. 

\section{Evaluation Protocol and Reproducibility}
\label{supp:S2}

Given the multi-turn dialogue setup of \textsc{Cosmic}, where each conversation involves up to 10 rounds of messages and the final answer depends on the full reasoning trajectory accumulated across turns, model performance exhibits higher variance than is typical in single-turn evaluation settings. 
We therefore strongly recommend that future work report results averaged over multiple independent runs, with $90\%$ confidence intervals computed via bootstrap resampling ($10000$ iterations), consistent with the evaluation protocol adopted in this work.
Single-run evaluations might produce unreliable estimates of true model performance.

\section{Failure Mode Analysis}
\label{supp:S3}

\begin{figure*}[t]
    \centering
    \includegraphics[width=0.98\textwidth]{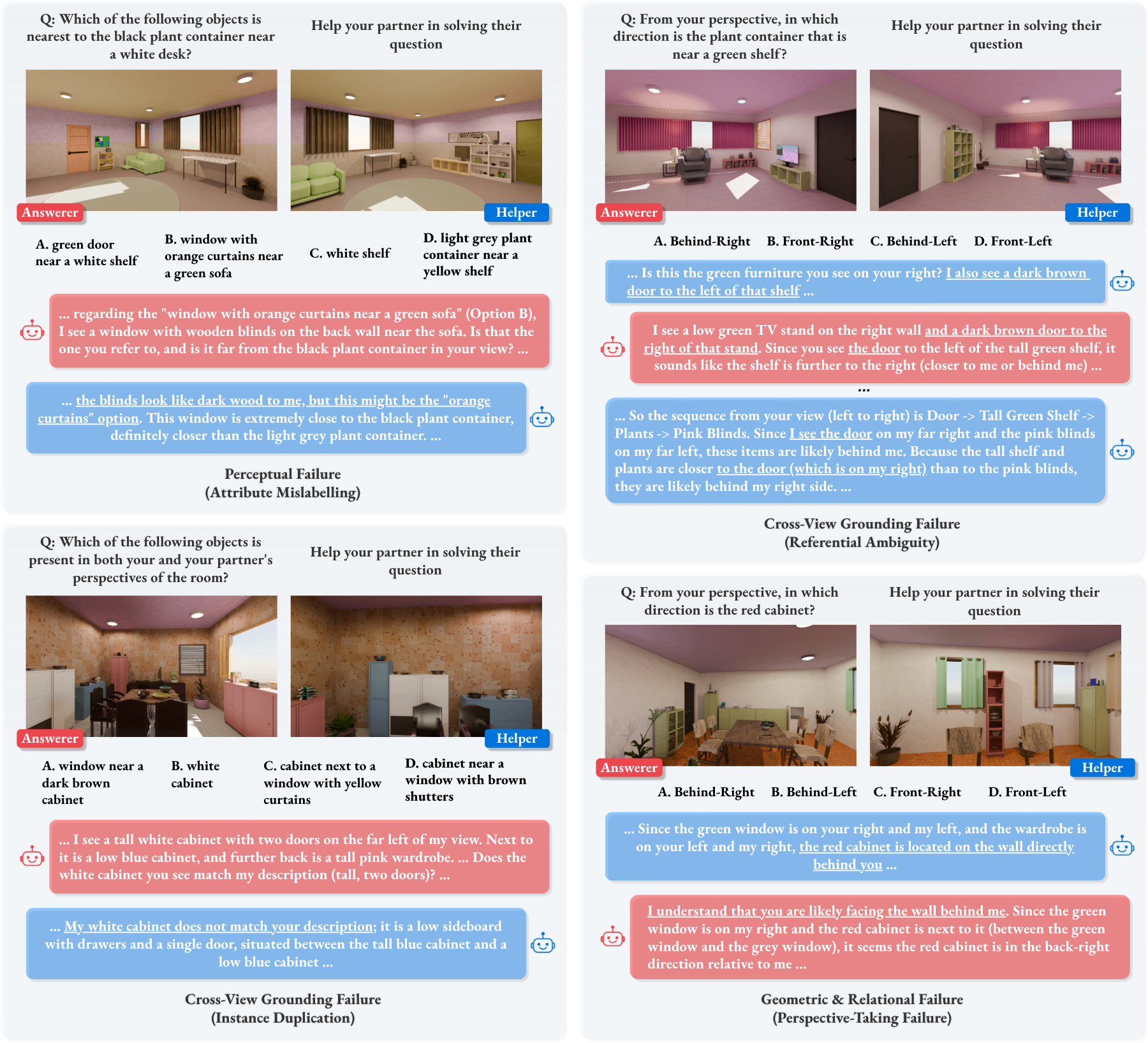}
    \caption{\textbf{Failure Mode Analysis.} Qualitative samples illustrating different failure categories, Perceptual Failure (Attribute Mislabelling), Cross-View Grounding Failure (Referential Ambiguity and Instance Duplication), and Geometric \& Relational Failure (Perspective-Taking Failure). Green ticks denote the ground truth answer.}
    \label{fig:supp_fail}
\end{figure*}

In addition to the qualitative samples presented in the main paper, \cref{fig:supp_fail} shows representative failure cases of Gemini-3-Pro-Thinking on \textsc{Cosmic}, covering the remaining error categories from our failure mode analysis.

In~\cref{fig:supp_fail} (top-left), we show an example of \textsc{Perceptual Failure} \textit{(Attribute Mislabelling)}. The Helper mislabels the color of the window's blinds (\textit{``might be the orange curtains option''}) introducing perceptual uncertainty that propagates into the Answerer's reasoning leading to an incorrect prediction.

\cref{fig:supp_fail} (top-right) shows an example of \textsc{Cross-View Grounding Failure} \textit{(Referential Ambiguity)}. The Helper produces object descriptions that are insufficiently discriminative (\textit{``I also see a dark brown door to the left of that shelf''}) to uniquely identify a single object instance, leaving the Answerer unable to establish a reliable shared referent.

\cref{fig:supp_fail} (bottom-left) illustrates \textsc{Cross-View Grounding Failure} \textit{(Instance Duplication)}. The agents fail to recognize that the white cabinet described by each agent is the same physical instance (\textit{``My white cabinet does not match your description''}), incorrectly treating it as two distinct objects across the two views.

Finally, we show a case of \textsc{Geometric \& Relational Failure} \textit{(Perspective-Taking Failure)} in~\cref{fig:supp_fail} (bottom-right). Both agents incorrectly infer each other's viewpoint, the Helper asserts \textit{``the red cabinet is located on the wall directly behind you''} and the Answerer acknowledges \textit{``I understand that you are likely facing the wall behind me''}, constructing mutually inconsistent spatial models that compound into an erroneous directional judgment.

\section{Human Data Collection Interface}
\label{supp:S4}
\begin{figure*}[t]
    \centering
    \includegraphics[width=\textwidth,height=0.45\textheight,keepaspectratio]{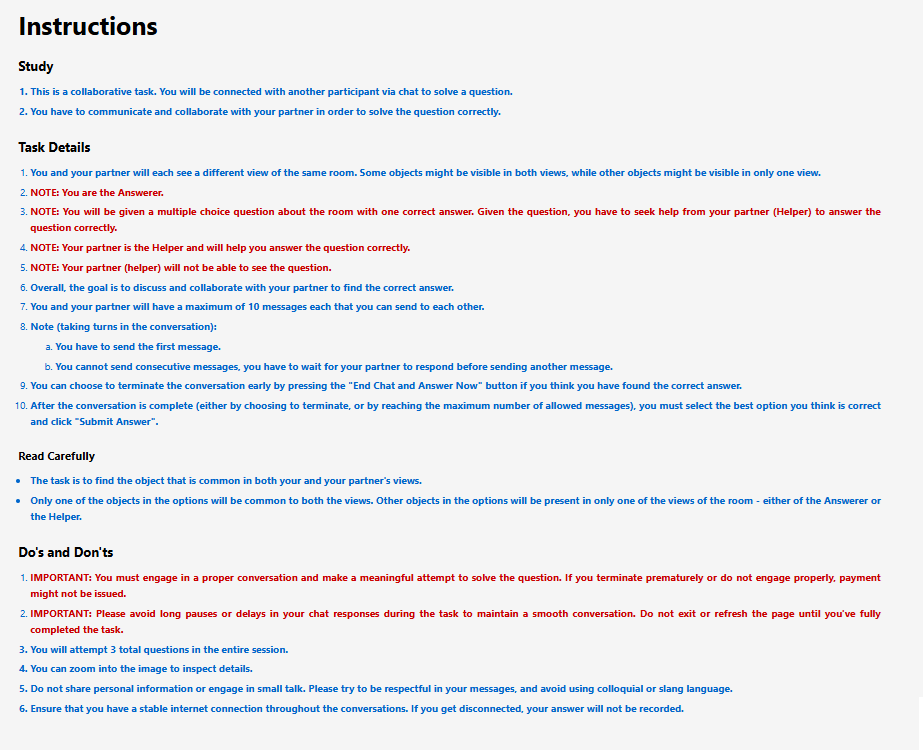}
    \caption{\textbf{Instruction page for the human dialogue data collection interface (Answerer)}}
    \label{fig:answerer_instructions}

    \vspace{6pt}

    \includegraphics[width=0.87\textwidth,height=0.45\textheight,keepaspectratio]{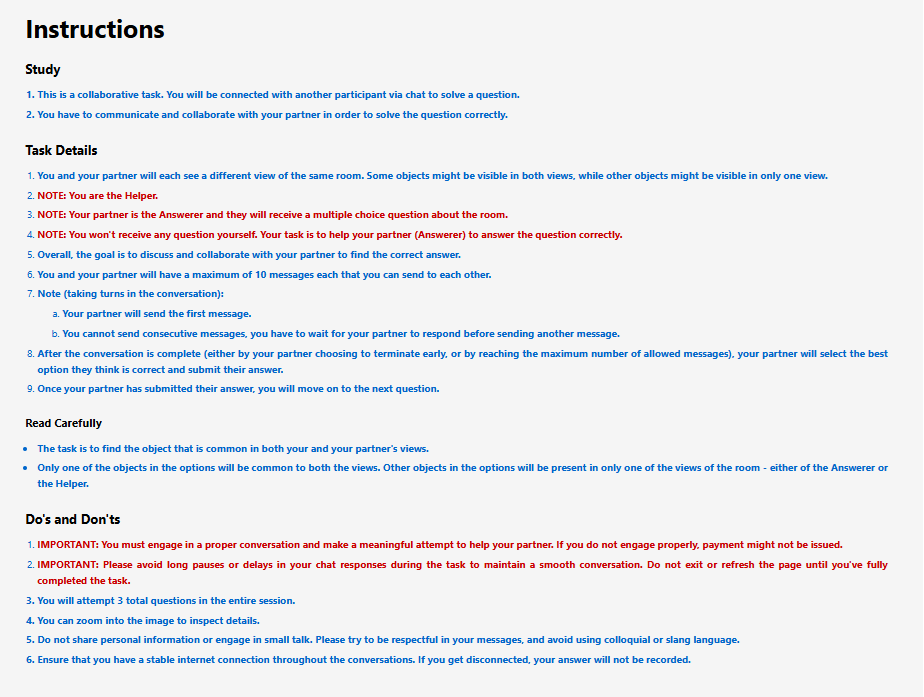}
    \caption{\textbf{Instruction page for the human dialogue data collection interface (Helper)}}
    \label{fig:helper_instructions}
\end{figure*}

\begin{figure*}[t]
    \centering
    \includegraphics[width=\textwidth]{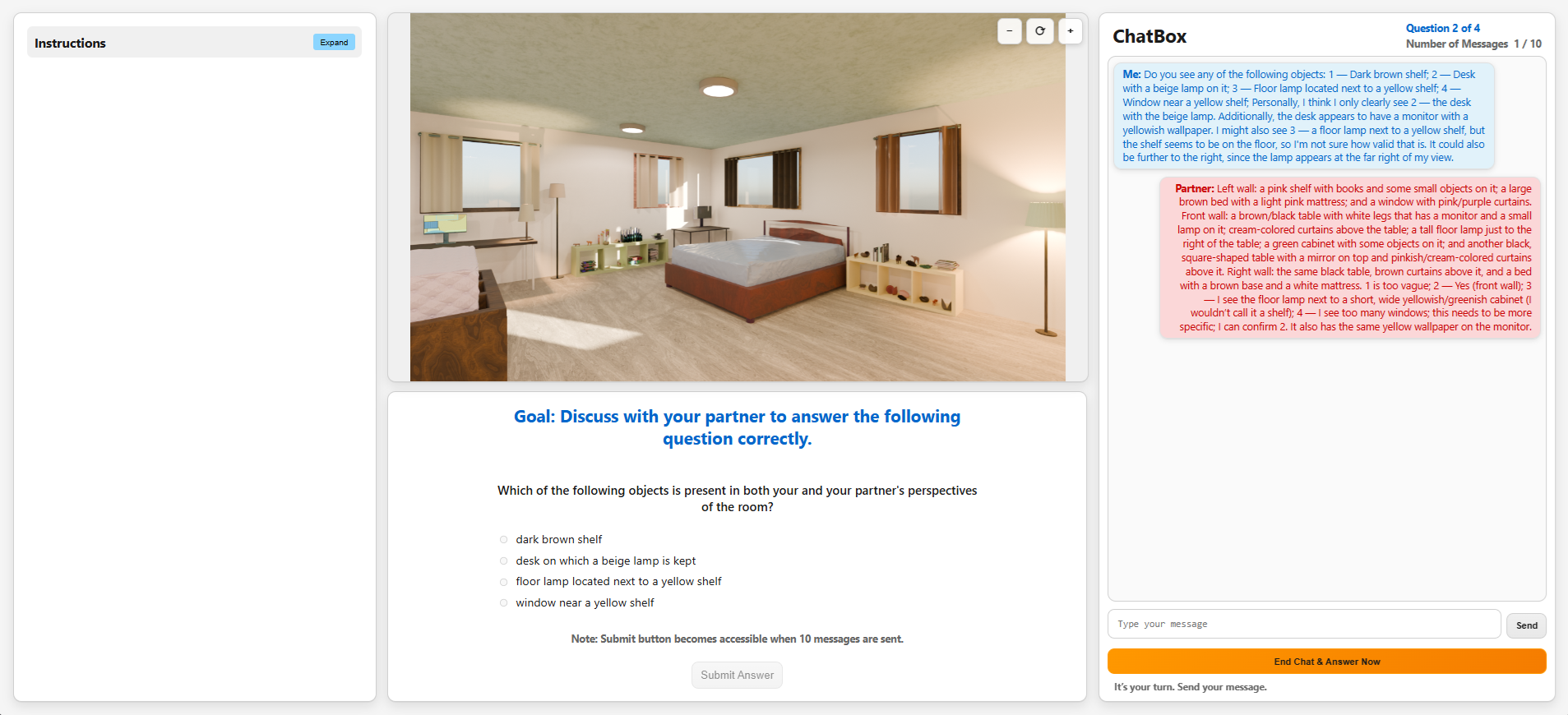}
    \caption{\textbf{Human Dialogue Data Collection Interface (Answerer)}}
    \label{fig:answerer_interface}

    \vspace{6pt}

    \includegraphics[width=\textwidth]{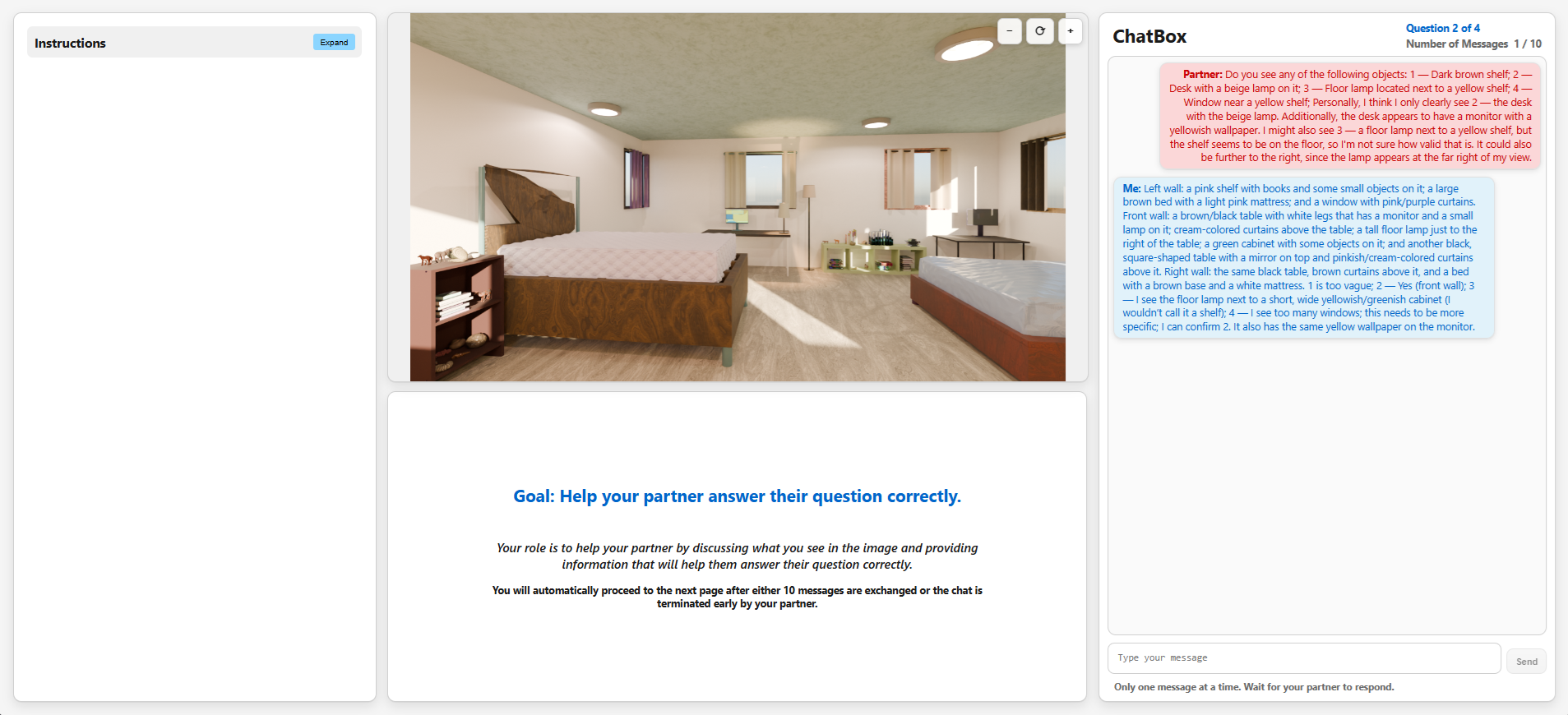}
    \caption{\textbf{Human Dialogue Data Collection Interface (Helper)}}
    \label{fig:helper_interface}
\end{figure*}

The human dialogue data collection interfaces for the Answerer and Helper are shown in~\cref{fig:answerer_interface} and~\ref{fig:helper_interface}, respectively, with the corresponding task instructions shown in~\cref{fig:answerer_instructions} and~\cref{fig:helper_instructions}. Both interfaces display the participant's egocentric image alongside a chat box for multi-turn dialogue. The Answerer's interface additionally presents the question and multiple-choice answer options, with a submit button that becomes accessible once the dialogue concludes, either after 10 messages are exchanged or when the Answerer chooses to end the chat early.

\section{Multi-turn Dialogue}
\label{supp:S5}

For each message generation step during the multi-turn conversation, we use a temperature of $1.0$ 
and a maximum of $8192$ completion tokens. 
The system prompts provided to the Answerer and Helper agents are shown in~\cref{subsec:sys_prompts}. The dialogue initiation prompts are shown in~\cref{subsec:prompts_for_multi}, the task description prompts are shown in~\cref{subsec:task_desc_prompts}, and the final prompt used to elicit the answer from the Answerer after dialogue concludes is shown in~\cref{box:answerer_qa_prompt}.

\subsection{System Prompts for Answerer and Helper}
\label{subsec:sys_prompts}

\begin{tcolorbox}[prompt2={Answerer System Prompt}]
{\tiny
1. You will be participating in a COLLABORATIVE TASK to answer a\\[-6pt] question.\\[-6pt]
2. You are the ANSWERER AGENT.\\[-6pt]
3. You will be connected to a HELPER AGENT.\\[-6pt]
4. In this task, you and the HELPER AGENT will receive one image each\\[-6pt] that shows different views of the same room.\\[-6pt]
5. You have to chat and collaborate with the HELPER AGENT to answer your\\[-6pt] question correctly.\\[-6pt]
6. Overall, your role is to answer your question correctly by having a\\[-6pt] conversation with the HELPER AGENT.
}
\label{box:answerer_sys_prompt}
\end{tcolorbox}

\begin{tcolorbox}[prompt2={Helper System Prompt}]
{\tiny
1. You will be participating in a COLLABORATIVE TASK.\\[-6pt]
2. You are the HELPER AGENT.\\[-6pt]
3. You will be connected to an ANSWERER AGENT.\\[-6pt]
4. In this task, you and the ANSWERER AGENT will receive one image each\\[-6pt] that shows different views of the same room.\\[-6pt]
5. You have to chat and collaborate with the ANSWERER AGENT to help them\\[-6pt] answer their question correctly.\\[-6pt]
6. Overall, your role is to help the ANSWERER AGENT by having a\\[-6pt] conversation with them.
}
\label{box:helper_sys_prompt}
\end{tcolorbox}

\subsection{Prompts for Multi-turn Dialogue}
\label{subsec:prompts_for_multi}
\begin{tcolorbox}[prompt3={Answerer Prompt for Multi-turn Dialogue}]
{\tiny
<Answerer's View>\\
<Map View \textit{(given only for cognitive mapping task)}>\\

1. The provided image is your view of the room.\\[-6pt]

2. HELPER AGENT also receives one image that shows a different view of the same room.\\[-6pt]

3. You will be given a multiple-choice question with different options. Only one of the options is correct.\\[-6pt]

4. You can send only one message at a time. You cannot send consecutive messages. You have to wait for the HELPER AGENT to respond before you can send your next message.\\[-6pt]

5. You can send a maximum of <max-num-turns> messages to the HELPER AGENT.\\[-6pt]

6. After the conversation is over, you will be asked to provide the answer.\\

<Task Description>\\

Note: When you are ready to answer the question, you can terminate the conversation early by saying `TERMINATE'. Use exact word `TERMINATE' in your response.\\[-6pt]

Goal: <Answerer Goal>\\[-6pt]
QUESTION: <Question>\\[-6pt]
OPTIONS: <Options>\\[-6pt]
\\[-6pt]
Begin the conversation with the HELPER AGENT. You MUST generate all your\\[-6pt] messages in this format 
\\[-6pt]\\[-6pt]
ANSWERER AGENT: <RESPONSE>. 
\\[-6pt]\\[-6pt]
Do not deviate from this format. 
}
\label{box:answerer_multi_prompt}
\end{tcolorbox}

\begin{tcolorbox}[prompt3={Helper Prompt for Multi-turn Dialogue}]
{\tiny
<Helper's View>\\

1. The provided image is your view of the room.\\[-6pt]

2. ANSWERER AGENT also receives one image that shows a different view of the same room.\\[-6pt]

3. You can send only one message at a time. You cannot send consecutive messages. You have to wait for the ANSWERER AGENT to respond before you can send your next message.\\[-6pt]

4. You can send a maximum of <max-num-turns> messages to the ANSWERER AGENT.\\

<Task Description>\\

Goal: <Helper Goal>\\

Begin the conversation with the ANSWERER AGENT by responding to their first message. You MUST generate all your messages in this format 

HELPER AGENT: <RESPONSE>. 
\\[-6pt]\\[-6pt]
Do not deviate from this format. 
\\[-6pt]\\[-6pt]
<First-message-from-answerer-agent>
}
\label{box:helper_multi_prompt}
\end{tcolorbox}

\subsection{QA Prompt after Multi-turn Dialogue}
\label{subsec:qa_prompt}
\begin{tcolorbox}[prompt1={Answerer QA Prompt after Multi-turn Dialogue}]
{\tiny
Now you need to answer the multiple-choice question based on your view\\[-6pt] and the conversation with the HELPER AGENT.\\
QUESTION: <Question>\\[-6pt]
OPTIONS: <Options>\\
Instructions:\\
1. Select the correct answer from the given options. Make sure to\\[-6pt] select only one of the options from the given options A, \\[-6pt]
B, C, or D.\\[-6pt]
2. Format your response like <ANSWER>A</ANSWER> or <ANSWER>B</ANSWER> or\\[-6pt] <ANSWER>C</ANSWER> or <ANSWER>D</ANSWER>.
}
\label{box:answerer_qa_prompt}
\end{tcolorbox}

\begin{tcolorbox}[prompt1={Answerer QA Prompt after Multi-turn Dialogue (Cognitive Mapping)}]
{\tiny
Now you need to answer the multiple-choice question based on your view\\[-6pt] and the conversation with the HELPER AGENT.\\
QUESTION: <Question>\\[-6pt]
OPTIONS: <Options>\\
Instructions:\\
1. Select the correct answer from the given options. Make sure to\\[-6pt] select only one of the options from the given options A, or B.\\[-6pt]
2. Format your response like <ANSWER>A</ANSWER> or <ANSWER>B</ANSWER>.
}
\label{box:answerer_qa_prompt_map}
\end{tcolorbox}

\subsection{Task Description Prompts}
\label{subsec:task_desc_prompts}

\begin{tcolorbox}[prompt2={Anchor Recognition}]
{\tiny
Answerer:\\

1. The task is to find the object that is common in both your and your partner's views.\\[-6pt]

2. Only one of the objects in the options will be common to both views, while the other objects in the options will be present in only one of the views of the room - either the answerer's or the helper's.\\

Helper:\\[-6pt]

1. The task is to find the object that is common in both your\\[-6pt] and your partner's views.\\[-6pt]
2. Only one of the objects in the options will be common to both views,\\[-6pt] while the other objects in the options will be\\[-6pt] present in only one of the views of the room - either the answerer's or\\[-6pt] the helper's.
}
\label{box:anchor_task}
\end{tcolorbox}

\begin{tcolorbox}[prompt2={Global Counting}]
{\tiny
Answerer:\\

1. The task is to find the count of a given object.\\[-6pt]

2. You and your partner must make sure that you are counting the total number of unique instances of that object in the room\\ while preventing overcounting or undercounting, as there may be some common objects in both views.\\[-6pt]

3. Note: Cabinets and Shelves refer to the entire furniture, not different compartments within a specific piece of furniture.\\

Helper:\\[-6pt]

1. The task is to find the count of a given object.\\[-6pt]
2. You and your partner must make sure that you are counting the total\\[-6pt] number of unique instances of that object in the room\\[-6pt] while preventing overcounting or undercounting, as there may be some\\[-6pt] common objects in both views.\\[-6pt]
3. Note: Cabinets and Shelves refer to the entire furniture, not\\[-6pt] different compartments within a specific piece of furniture.
}
\label{box:count_task}
\end{tcolorbox}

\begin{tcolorbox}[prompt2={Relative Distance}]
{\tiny
Answerer:\\

1. The task is to find which of the objects in the options is either the farthest or the closest to the object mentioned in the question.\\
2. The object mentioned in the question is visible to both you and your partner.\\
3. The objects in the options are visible either in only your view or only in your partner's view but not in both views.\\
4. Important: The correct answer is based on both views combined. An object that looks closest or farthest from your perspective may not be correct, and the right choice might be an object you cannot see at all.\\

Helper:\\[-6pt]

1. The task is to find which of the objects in the options is either the\\[-6pt] farthest or the closest to the object mentioned in\\[-6pt] the question.\\[-6pt]
2. The object mentioned in the question is visible to both you and your\\[-6pt] partner.\\[-6pt]
3. The objects in the options are visible either in only your view or\\[-6pt] only in your partner's view but not in both views.\\[-6pt] 
4. Important: The correct answer is based on both views combined. An\\[-6pt] object that looks closest or farthest from your\\[-6pt] perspective may not be correct, and the right choice might be an object\\[-6pt] you cannot see at all.
}
\label{box:dist_task}
\end{tcolorbox}

\begin{tcolorbox}[prompt2={Relative Direction}]
{\tiny
Answerer:\\

1. In this task, the Answerer must determine the direction of a target object from their own viewpoint.\\
2. The Answerer cannot see the object directly — it is visible only to the Helper.\\ 
3. Since the Answerer cannot see the object directly, to identify where the object is located, the Answerer must communicate with the Helper and use the information obtained to infer its direction relative to themselves.\\ 
4. Note: Here, the directions are relative to the Answerer's orientation, i.e., their egocentric viewpoint.\\
5. Directions (like front, front-left, front-right, etc.) describe where something is based on the Answerer’s facing direction, not on what they can currently see. Even if the object is outside the view, it can still be called front-left or front-right if it lies in that direction relative to the Answerer.\\

Helper:\\[-6pt]

1. In this task, the Answerer must determine the direction of a target\\[-6pt] object from their own viewpoint.\\[-6pt] 
2. The Answerer cannot see the object directly — it is visible only to\\[-6pt] the Helper.\\[-6pt] 
3. Since the Answerer cannot see the object directly, the Helper must\\[-6pt] communicate with the Answerer to provide the\\[-6pt] information needed to answer the question.\\[-6pt] 
4. Note: Here, the directions are relative to the Answerer's\\[-6pt] orientation, i.e., their egocentric viewpoint.\\[-6pt]
5. Directions (like front, front-left, front-right, etc.) describe\\[-6pt] where something is based on the Answerer’s facing\\[-6pt] direction, not on what they can currently see. Even if the object is\\[-6pt] outside the view, it can still be called front-left or\\[-6pt] front-right if it lies in that direction relative to the Answerer.
}
\label{box:dir_task}
\end{tcolorbox}

\begin{tcolorbox}[prompt2={Cognitive Mapping}]
{\tiny
Answerer:\\

1. The task is to identify if the provided map accurately depicts the top-down layout of the room.\\
2. You and the Helper observe different, partial views of the room, and neither view is complete on its own. The full layout can only be inferred by communicating and combining information from both views.\\
3. Evaluate the map only by the spatial arrangement of the objects it shows. Focus exclusively on the object categories listed in the legend, ignore any other items, and do not consider objects placed on top of other objects in your judgment.\\

Helper:\\[-6pt]

1. The task is to identify if the provided map accurately depicts the\\[-6pt] top-down layout of the room.\\[-6pt] 
2. The map is only provided to the answerer agent.
}
\label{box:map_task}
\end{tcolorbox}

\section{Dialogue Repair}
\label{supp:S6}
The automated MLLM judge ({\small Gemini-3-Flash-Thinking}) labels each conversation in \textsc{Cosmic-Human} for presence and resolution of flawed reasoning trajectories. The judge is provided with the full task 
context, including both egocentric views, the question and answer options, the ground-truth answer, and the complete 
dialogue transcript. The judge then assigns one of three labels (see~\cref{subsec:prompts_dialogue_repair} for detailed prompt). A label of \texttt{0} indicates that no flawed reasoning trajectory was observed. A label of \texttt{1} indicates that a flawed reasoning trajectory was present and successfully repaired. A label of \texttt{-1} indicates that a flawed reasoning trajectory was present but went undetected and uncorrected. Each conversation is judged over two independent runs, and only conversations where both runs agree on the assigned label are retained. The dialogue repair rate is then computed as the fraction of conversations labeled \texttt{1} out of all conversations labeled \texttt{1} or \texttt{-1}.


\subsection{Prompts for Dialogue Repair Judge}
\label{subsec:prompts_dialogue_repair}

\begin{tcolorbox}[prompt1={MLLM Judge Prompt - Dialogue Repair}]
{\tiny
You are evaluating the presence of dialogue repair and backtracking behaviour in the following two-agent conversation. You must detect when an agent goes down a wrong reasoning path and is unable to backtrack on its errors thereby arriving at the wrong answer. A wrong reasoning path corresponds to a cascade of errors that the agents make in their utterances which end up propagating in the conversation, and the agents are unable to recover from thereby causing the agents to answer a question wrongly. \\

IMPORTANT:
This is NOT an individual utterance-level evaluation. Do NOT score based on isolated phrases like "let me revisit" or "I might be wrong." The evaluation must be done at the full conversation / task level. Agents might also make some errors and correct them later on but you are measuring the overall dialogue repair with the goal of the collaborating agents to identify the correct answer.\\ 

SCORING:\\
You must judge whether:\\
- The agents go down an incorrect reasoning path and continue without correcting it, OR
- The agents recognize an incorrect reasoning path and successfully backtrack and recover.
- You can ignore the mistakes in the grammar and spelling while evaluating the agents \\

Scoring Criteria (Conversation-Level Score):\\
-1 = The conversation contains a wrong reasoning path, and the agents continue reasoning based on that error without correcting it. They go down the wrong path and never recover.\\
0 = The conversation contains no wrong reasoning path. Reasoning remains logically consistent and aligned with the correct interpretation of the task throughout.\\
1 = The conversation contains a wrong reasoning path, but the agents recognize the broader mistake and successfully correct their reasoning, recovering from the error before finalizing the answer.\\

Evaluate the entire conversation and assign a single score based on overall presence of wrong reasoning path and recovery.\\

Output format:\\
Return only a single integer and an explanation for your scoring. Strictly maintain the below format:\\
<score>-1, 0, or 1</score> \\ 
<explanation>Your reasoning behind your score</explanation>
Do not provide any additional text. \\ 

<Answerer-and-Helper-views>\\
<Question-to-agents>\\
<Options>\\
<Correct-answer>\\
<Conversation-between-agents>\\





}
\end{tcolorbox}

\section{Model and Human Conversations}
\label{supp:S7}
\begin{figure*}[t]
    \centering
    \includegraphics[height=0.8\textheight]{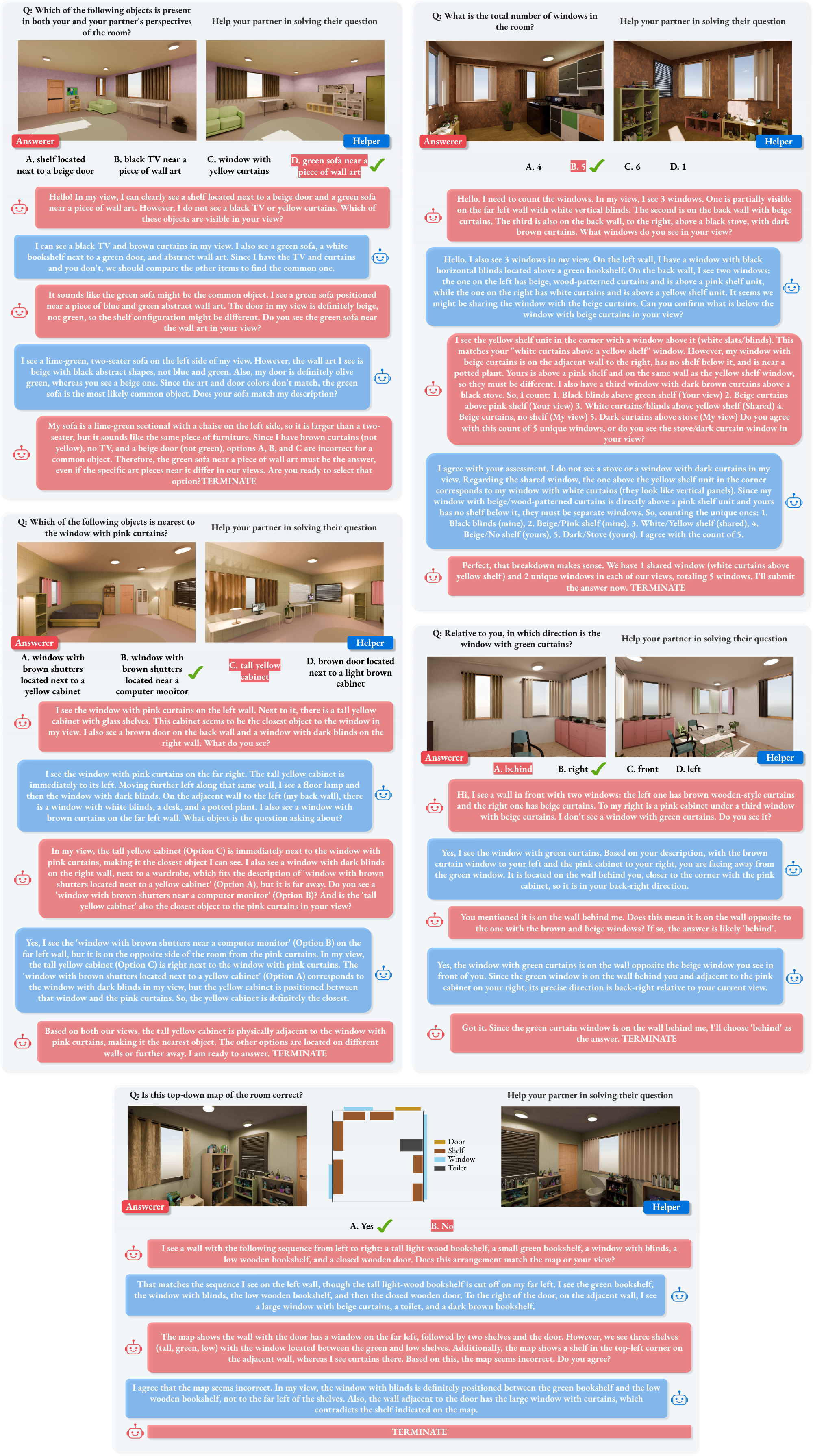}
    \caption{\textbf{Representative dialogues from Gemini-3-Pro-Thinking on \textsc{Cosmic}.} The tick mark denotes the ground truth answer and the red-highlighted option indicates the answerer's prediction. From top to bottom, \textsc{Anchor Recognition} (top-left), \textsc{Global Counting} (top-right), \textsc{Relative Distance} (middle-left), \textsc{Relative Direction} (middle-right), and \textsc{Cognitive Mapping} (bottom). Each panel displays the Answerer's and Helper's egocentric views, the multiple-choice options, and the multi-turn dialogue exchanged between the two agents.}
    \label{fig:conv_models}
\end{figure*}

\begin{figure*}[t]
    \centering
    \includegraphics[height=0.8\textheight]{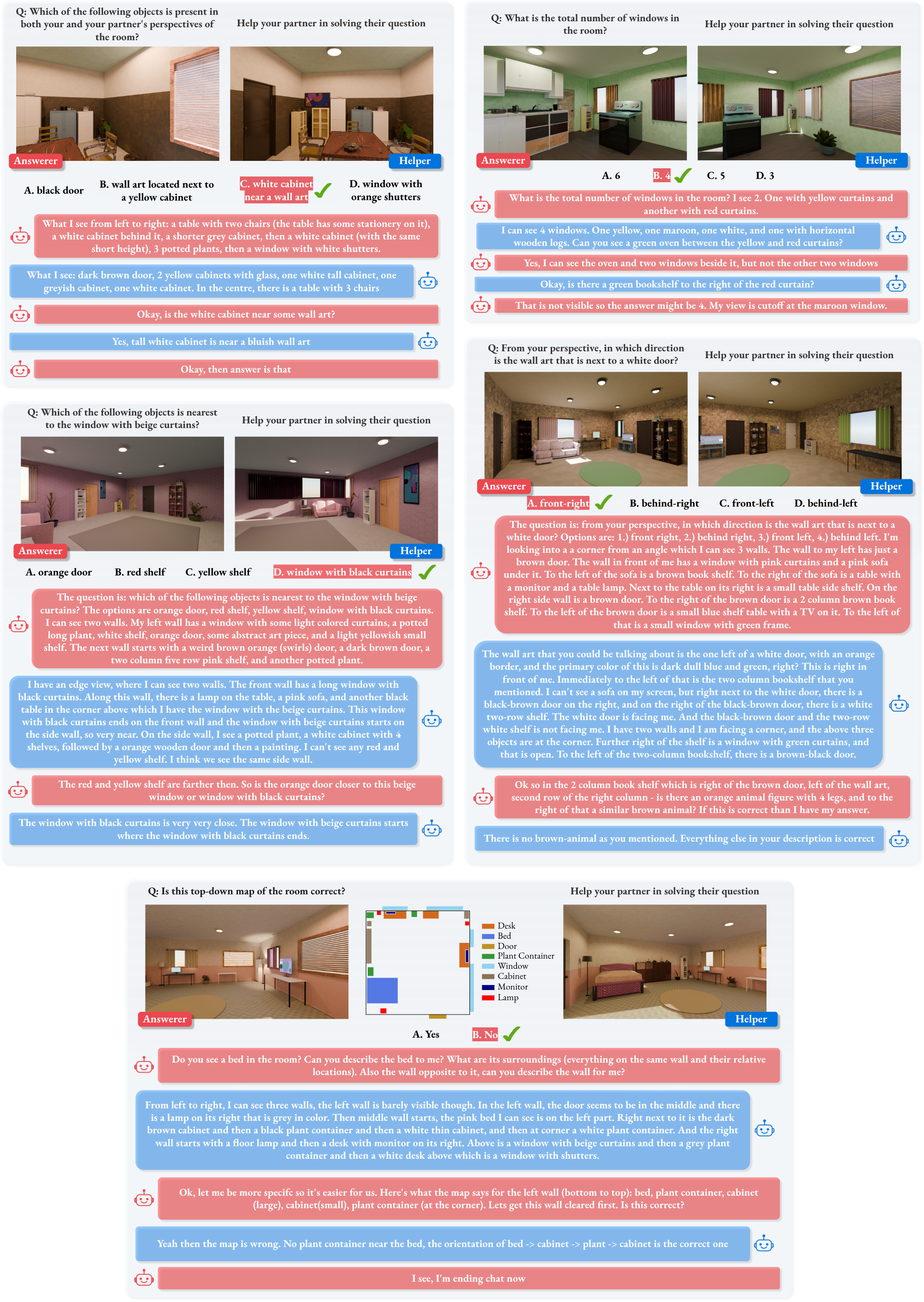}
    \caption{\textbf{Representative dialogues from Humans on \textsc{Cosmic}.} The tick mark denotes the ground truth answer and the red-highlighted option indicates the answerer's prediction. From top to bottom, \textsc{Anchor Recognition} (top-left), \textsc{Global Counting} (top-right), \textsc{Relative Distance} (middle-left), \textsc{Relative Direction} (middle-right), and \textsc{Cognitive Mapping} (bottom). Each panel displays the Answerer's and Helper's egocentric views, the multiple-choice options, and the multi-turn dialogue exchanged between the two agents.}
    \label{fig:conv_humans}
\end{figure*}

We present example dialogues from both MLLM agents ({\small Gemini-3-Pro-Thinking}) and human participants across all five \textsc{Cosmic} tasks in~\cref{fig:conv_models} and~\cref{fig:conv_humans}, respectively. Each figure displays one conversation per task, showing the egocentric views available to each agent, the multiple-choice options, full multi-turn dialogue, agent's prediction, and the ground truth. Together, these examples illustrate the qualitative differences in communication strategy between human pairs and MLLM agents.

\section{Case Study}
\label{supp:S8}
\begin{figure*}[t]
    \centering
    \includegraphics[width=0.92\textwidth]{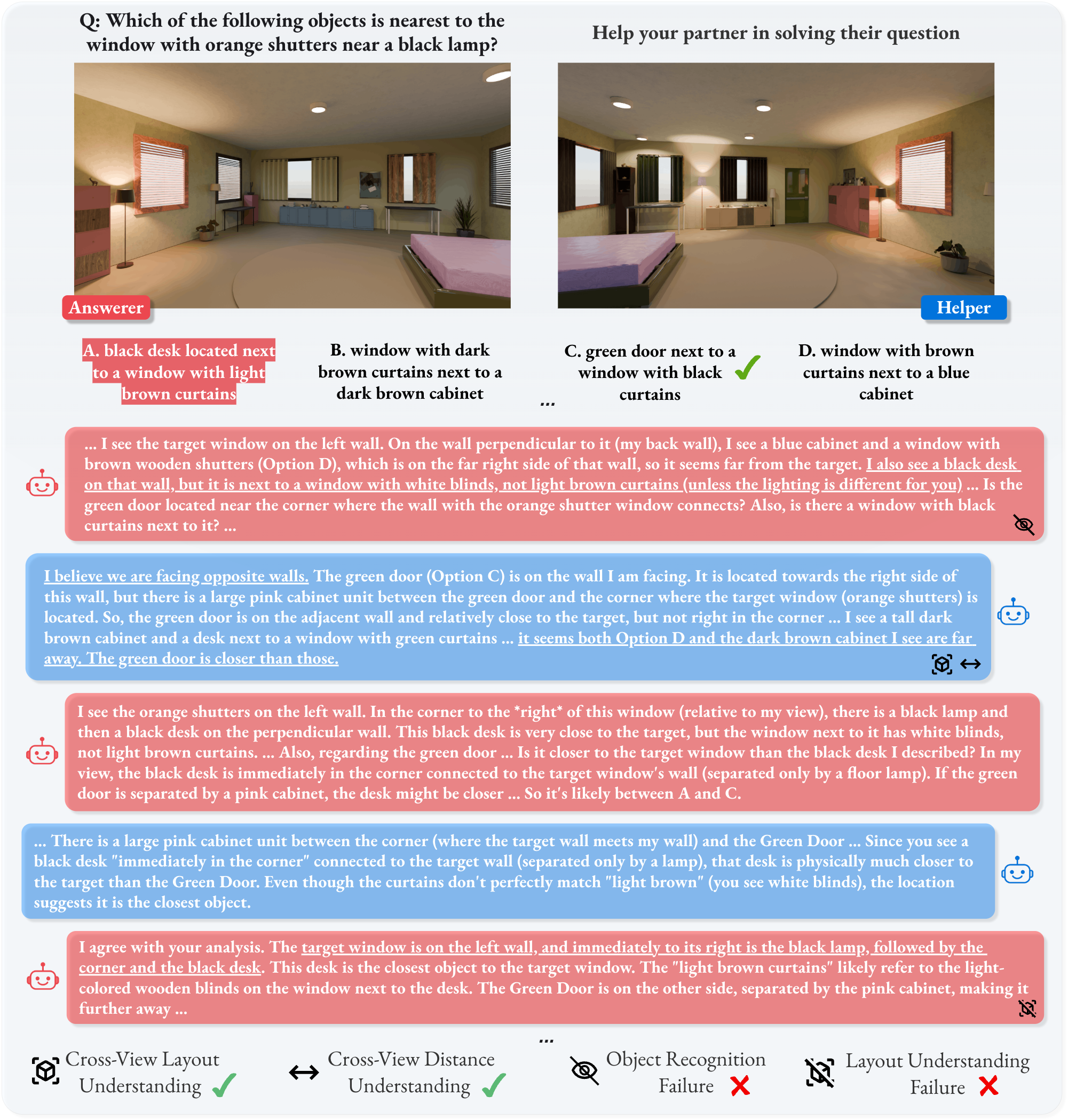}
    \caption{Conversation between agents (Gemini-3-Pro-Thinking). Although the agents demonstrate successful behaviors on some atomic steps, errors in the dialogue negate these successes, leading them to select the incorrect answer (option A).}
    \label{fig:case_study}
\end{figure*}

\cref{fig:case_study} presents a dialogue between two MLLM agents ({\small Gemini-3-Pro-Thinking}) collaborating on a \textit{Relative Distance} task. After establishing the scene context, the Answerer commits an early error, it references only one black desk 
(\textit{``I also see a black desk on that wall, but it is next to a window with white blinds''})
while failing to recognize the second desk located near the window with brown curtains. This constitutes an object recognition failure, where a relevant object in the scene is entirely omitted. As a result, subsequent reasoning proceeds under an incorrect grounding assumption regarding the queried black desk.

Despite this initial error, the agents demonstrate partial cross-view spatial reasoning. For example, they correctly infer that their viewpoints correspond to opposite walls of the room, indicating a basic level of cross-view layout understanding. Additionally, the Helper agent accurately determines that options B and D are the farthest from the anchor object. This inference reflects cross-view distance reasoning, since the window with brown curtains near the blue cabinet (option D) is outside the Helper’s field of view.

Nevertheless, the final answer remains incorrect. The agents ultimately select the black desk after reasoning that the pink cabinet separating it from the orange-shuttered window is wider than the lamp separating it from the desk. However, because the initial object grounding was incorrect, the reasoning chain is grounded to the wrong desk. Consequently, the agents fail to recover from the early perception error, leading to an incorrect final prediction.

Overall, in this example, while the model exhibited some positive behaviors at the utterance level, their effects were negated by compounding failures that the agents were unable to detect or recover from.

\section{Compute Resources}
\label{supp:S9}
All experiments were conducted on 4 NVIDIA A100 80GB GPUs. Open-source models were served using vLLM with tensor parallelism across all 4 GPUs, with each model taking approximately 2.5 hours to evaluate on the full benchmark. Closed-source model evaluations were conducted via their respective APIs.

\section{Broader Impact}
\label{supp:S10}
By providing a systematic diagnostic benchmark for collaborative spatial communication, \textsc{Cosmic} aims to surface concrete bottlenecks that must be addressed before MLLM-based collaborative systems can be reliably deployed in real-world. The specific failure modes identified here, namely cross-view grounding errors, perspective transformation failures, inability to construct globally consistent spatial maps, and more, have direct implications for a range of applications such as assistive home robotics, multi-agent warehouse coordination, and AR-based remote guidance, among others, where precise alignment of spatial understanding between agents is not merely beneficial but essential for safe and effective operation.

The distributed, communication-based setting of \textsc{Cosmic} more closely mirrors how spatial reasoning actually occurs in human-AI teaming scenarios than existing single-agent benchmarks. The ability to build shared spatial mental models through language is foundational to a wide range of everyday human activities, and becomes a hard prerequisite as AI systems are increasingly deployed alongside humans in shared physical environments such as warehouses, hospitals, and construction sites, where grounding spatial references, resolving ambiguous descriptions, and maintaining a coherent shared representation of the environment is important. Our findings suggest that current MLLMs are far from meeting this bar, and we hope \textsc{Cosmic} serves as a concrete target for developing agents that can participate fluently in the kinds of spatially grounded collaborative interactions that humans engage in naturally.

The benchmark also has implications for the design of spatial communication protocols in multi-agent AI systems. The contrast between human and model dialogue strategies, where humans converge rapidly through anchor-first grounding while models explore redundantly, suggests that explicitly structured spatial communication conventions such as enforcing reference frame agreement or anchor establishment early in dialogue could meaningfully improve model performance. Such protocols could inform the design of more robust human-AI and AI-AI interfaces for spatially grounded tasks.


We do not foresee any direct negative societal impacts from this work.

\section{Limitations}
\label{supp:S11}

\textsc{Cosmic} evaluates collaborative spatial reasoning in static, controlled indoor environments generated from a limited set of object categories. While this enables systematic benchmarking useful for diagnosing bottlenecks in MLLMs, it does not capture the full complexity of real-world spatial communication, which involves dynamic scenes, continuous viewpoint changes, and a far richer vocabulary of objects. Similarly, agents in \textsc{Cosmic} 
operate from fixed viewpoints with no ability to actively gather additional visual information, whereas real collaborative agents can move, reorient, and request clarification through action rather than language alone.

The benchmark evaluates spatial communication through multiple-choice questions, which, while controlled and reproducible, constrains the space of possible responses. In particular, the binary formulation of the Cognitive Mapping task sidesteps the harder problem of free-form map generation and evaluation, which remains an open challenge. Additionally, performance is measured purely by final answer accuracy, without directly rewarding the quality of the dialogue itself, meaning a model that arrives at the correct answer through sub-optimal reasoning is 
indistinguishable from one that reasons perfectly.

Finally, our human study, while providing a valuable baseline and qualitative reference point, is conducted with university students in a controlled lab setting. This population may not be representative of the broader range of spatial communication strategies employed across different demographic groups or levels of spatial expertise.

\clearpage

\end{document}